\documentclass[11pt,a4paper]{article}
\usepackage[hyperref]{acl2020}
\usepackage{times}
\usepackage{latexsym}
\usepackage{amsmath}
\usepackage{amssymb}
\usepackage{dialogue}
\usepackage{graphicx}
\usepackage{tabularx}
\usepackage[normalem]{ulem}
\usepackage{xcolor}
\usepackage{tikz}
\usepackage{float}
\usepackage{enumitem}

\useunder{\uline}{\ul}{}

\usepackage{microtype}

\aclfinalcopy 

\title{Continual Learning in Task-Oriented Dialogue Systems}

\author{Andrea Madotto\thanks{\quad Work done during internship at Facebook}$^*$, Zhaojiang Lin$^*$, Zhenpeng Zhou$^\dagger$, Seungwhan Moon$^\dagger$,\\ \textbf{Paul Crook$^\dagger$, Bing Liu$^\dagger$, Zhou Yu$^\dagger$, Eunjoon Cho$^\dagger$, Zhiguang Wang$^\dagger$}\\
  $^*$The Hong Kong University of Science and Technology \\
  $^\dagger$Facebook \\
  \{\texttt{amadotto,zlinao}\}\texttt{@connect.ust.hk}\\
  \{\texttt{zzp,shanemoon,pacrook,bingl,zhoujoyu,zgwang}\}\texttt{@fb.com}\\
  }

\date{}

\begin{document}
\maketitle
\begin{abstract}
Continual learning in task-oriented dialogue systems can allow us to add new domains and functionalities through time without incurring the high cost of a whole system retraining. In this paper, we propose a continual learning benchmark for task-oriented dialogue systems with 37 domains to be learned continuously in four settings, such as intent recognition, state tracking, natural language generation, and end-to-end. Moreover, we implement and compare multiple existing continual learning baselines, and we propose a simple yet effective architectural method based on residual adapters. Our experiments demonstrate that the proposed architectural method and a simple replay-based strategy perform comparably well but they both achieve inferior performance to the multi-task learning baseline, in where all the data are shown at once, showing that continual learning in task-oriented dialogue systems is a challenging task. Furthermore, we reveal several trade-off between different continual learning methods in term of parameter usage and memory size, which are important in the design of a task-oriented dialogue system. The proposed benchmark is released together with several baselines to promote more research in this direction. 
\end{abstract}


\section{Introduction}
Task-oriented dialogue systems (ToDs) are the core technology of the current state-of-the-art smart assistant (e.g. Alexa, Siri, Portal etc.). These systems are either modularized, Natural Language Understanding (NLU), Dialogue State Tracking (DST), Dialogue Policy (DP) and Natural Language Generation (NLG), or end-to-end, where a single model implicitly learn how to issue APIs (i.e., NLU+DST) and system responses (i.e., NLG). 

\begin{figure}[t]
    \centering
    \includegraphics[width=\linewidth]{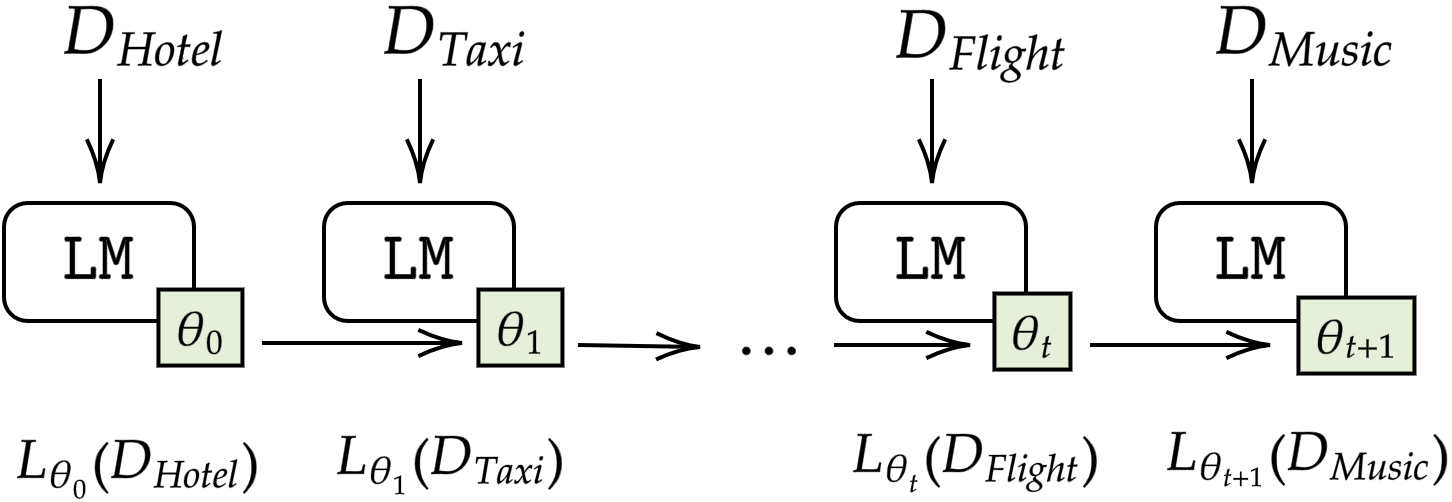}
    \caption{In Continual Learning, the model is trained one dataset at the time. In this instance, the model is first trained on data from the hotel domain $D_{Hotel}$ then on the Taxi $D_{Taxi}$ and so on. The parameter of the model is updated sequentially based on the loss function $L$. }
    \label{fig:example}
\end{figure}

These systems are often updated with new features based on the user needs, e.g., adding new slots and intents, or even completely new domains. However, existing dialogue models are trained with the assumption of having a fixed dataset at the beginning of the training, and they are not designed to add new domains and functionalities through time, without incurring the high cost of a whole system retraining. Therefore, the ability to acquire new knowledge continuously, a.k.a. Continual Learning (CL)~\cite{thrun2012learning}, is crucial in the design of a dialogue system. Figure~\ref{fig:example} shows an high-level intuition of CL in ToDs.  

In this setting the main challenge is catastrophic forgetting~\cite{mccloskey1989catastrophic}. This phenomena happens since there is a distributional shift between the tasks in the curriculum which leads to catastrophic forgetting the previously acquired knowledge. To overcome this challenge three kind of methods are usually deployed such as loss \textit{regularization}, for avoiding to interfere with the previously learned task, \textit{rehearsal}, which uses an episodic memory to recall previously learned tasks, and \textit{architectural}, which adds task-specific parameters for each learned task. However, architectural methods are usually not considered as a baseline, especially in sequence-to-sequence generation tasks~\cite{sun2019lamol}, because they usually \textit{require a further step} during testing for selecting which parameter to use for the given task.  

To the best our knowledge, Continual Learning (CL) in task-oriented dialogue systems~\cite{lee2017toward} is mostly unexplored or it has been studied in specific settings (e.g. NLG~\cite{mi2020continual}) using only few tasks learned continuously. Given the importance of the task in the dialogue setting, we believe that a more comprehensive investigation is required, especially by comparing multiple settings and baselines. Therefore in this paper:
\begin{enumerate}[noitemsep]
    \item we propose a benchmark for Continual Learning in ToDs, with 37 tasks to be learned continuously on 4 settings,
    \item we propose a simple yet effective architectural CL method based on residual adapters~\cite{houlsby2019parameter} that can continuously learn tasks without the need of a task classifier a testing time, and
    \item we analyse the trade-off between performance, number-of-parameters, and episodic memories sizes of the three main categories of CL methods (i.e., Regularization, Rehearsal, Architectural).
\end{enumerate}  

In Section~\ref{sec:back} we introduce the basic concepts and notation used throughout the paper, for both task-oriented dialogue modelling and continual learning, in Section~\ref{sec:adpt} we introduce the proposed architectural CL method, in Section~\ref{sec:experiment} we describe datasets, baselines, evaluation metrics and experimental settings, and Section~\ref{sec:results} we describe the main findings of the paper.   

\section{Background}\label{sec:back}

\begin{figure}
    \centering
    \includegraphics[width=\linewidth]{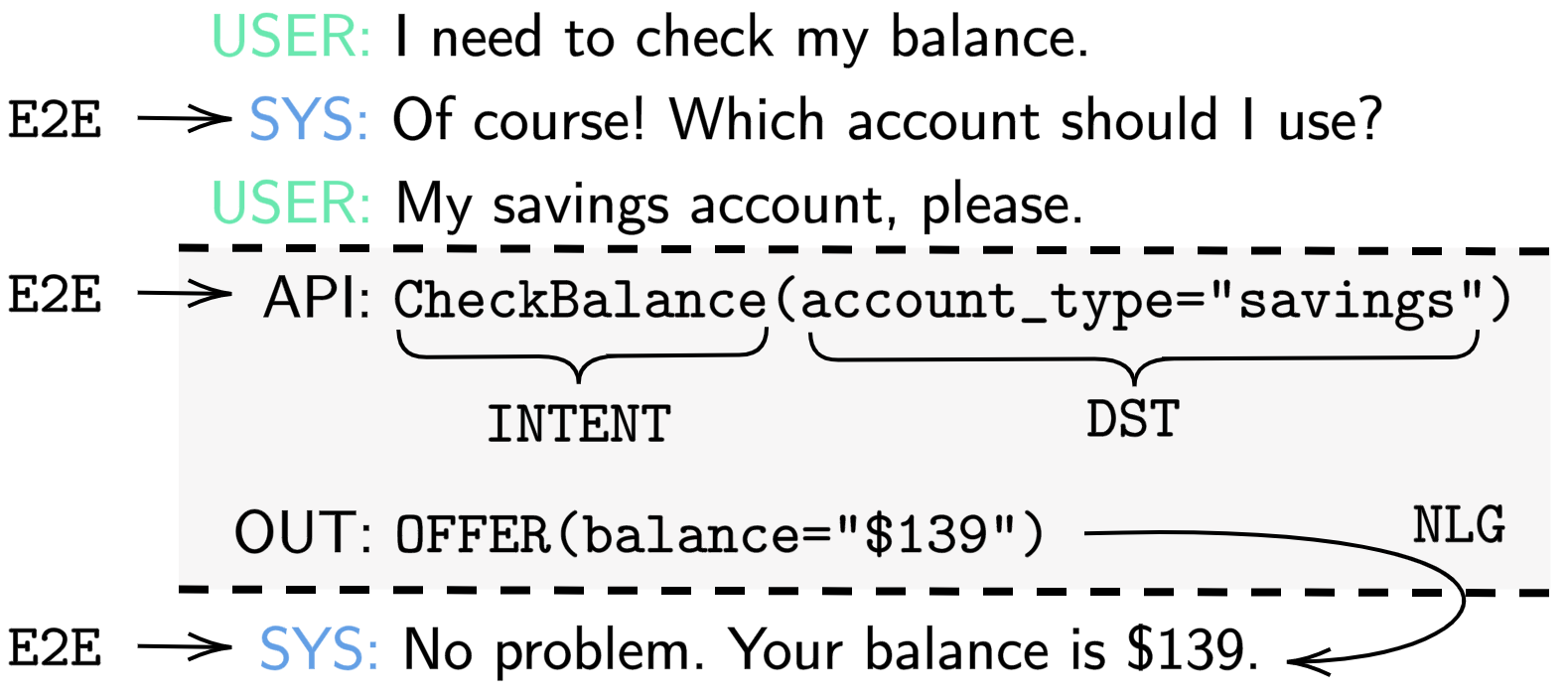}
    \caption{Example of input-out pairs, for the four settings, INTENT, DST, NLG and end-to-end (E2E). }
    \label{fig:example1}
\end{figure}

\subsection{Task-Oriented Dialogue Modelling} \label{subsec:preproc}
In this paper, we model task-oriented dialogue systems as a seq2seq generation task~\cite{lei2018sequicity,byrne2020tickettalk} that generates both api-calls and responses. As shown in Figure~\ref{fig:example1}, the model uses the dialogue history to generate an api-call, which is the concatenation of the user intent plus the current dialogue state and then uses its return, which can be empty or the system speech-act, to generate the system response. This modelling choice is guided by the existing annotated dialogue datasets, which provide the intent and the dialogue state of the user at every turn, and the speech-act of the system; and it allows us to defined four distinct settings for studying continual learning such as Intent Recognition (INTENT), Dialogue State Tracking (DST), Natural Language Generation (NLG) and end-to-end (E2E). In the coming paragraphs, we will formally describe the four setting as different input-out pairs for a seq2seq model. 
\paragraph{Data-Formatting} Let us define the dialogue history $H$ as a single sequence of tokens from the concatenation of the alternating utterances from the user and the system turns respectively. Without loss of generality, we assume that $H$ has all the dialogue history without the last system utterance denoted as $S$. For distinguishing between speakers, we add two special tokens at the beginning of every utterance i.e., \texttt{USER:} for the user utterance and \texttt{SYSTEM:} for the system utterance. Then, we define an api-call, denoted by $S_{API}$, as the concatenation of the api-name, i.e., the user-intent, and its arguments, i.e., slot-value pairs from the DST. The following syntax is used:  
\begin{equation}
X_{API} = \underbrace{\mathbf{I}}_{\text{Intent}} \ (\underbrace{s_{1}=v_{1}, \ldots,s_{k}=v_{p}}_{\text {Slot-value pairs}}) \label{API}
\end{equation}
where $\mathbf{I}$ is an intent or the api-name, $s_i$ the slot-name and $v_i$ one of the possible values for the slot $s_i$. The return of the api-call is either an empty string, thus the model uses the dialogue history to generate a response, or a speech-act, denoted as $S_{OUT}$, in the same format as the api-call in Equation~\ref{API}. Similar to the dialogue history, we define two special tokens \texttt{API:} and \texttt{OUT:} for triggering the model to generate the api-call and for distinguishing the return of the api from the dialogue history respectively. Based on this pre-processing, we define the four settings used in this paper.

Without loss of generality, we define the three modularized setting by their input-out pairs: 

\[
\begin{array}{cc}
H \rightarrow \mathbf{I} & \mathrm{(INTENT)} \\
H \rightarrow \mathbf{I}(s_{1}=v_{1}, \ldots,s_{k}=v_{p}) & \mathrm{(DST)} \\
\underbrace{\mathbf{I}(s_{1}=v_{1}, \ldots,s_{k}=v_{p})}_{S_{OUT}}\rightarrow S & \mathrm{(NLG)}
\end{array}
\]
where instead the end-to-end (E2E) setting as the pairs:
\[
\begin{array}{c}
 H \rightarrow \underbrace{\mathbf{I}(s_{1}=v_{1}, \ldots,s_{k}=v_{p})}_{S_{API}} \\
H + \underbrace{\mathbf{I}(s_{1}=v_{1}, \ldots,s_{k}=v_{p})}_{S_{OUT}}\rightarrow S
\end{array}
\]
Often, $S_{OUT}$ is empty and thus the model is though to map the dialogue history to the response directly ($H \rightarrow S$). An example of input-out pairs is shown in Figure~\ref{fig:example1}. 

Finally, we define a dialogue dataset as $D_K=\{(X_i,Y_i)\}_{i}^{n}$, where $(X_i,Y_i)$ is a general input-out pair from one of the four settings in consideration, and $K$ the dialogue domain under consideration (e.g., Hotel).   

\paragraph{Model} In this paper, we employ a decoder-only Language Models (e.g. GPT-2), which is inline with the current state-of-the-art task-oriented dialogue models \citet{peng2020soloist,hosseini2020simple}. Then, given the concatenation of the input $X= \{x_{0},\dots,x_{n}\}$ and output $Y= \{x_{n+1},\dots,x_{m}\}$ sequences, we compute the conditional language model distribution using the chain rule of probability~\cite{bengio2003neural} as:
\begin{equation}
    p_{\theta}(X|Y) =  \prod_{i=0}^{n+m}p_{\theta}(x_i|x_{0}, \cdots, x_{i-1}).
\end{equation}
where $\theta$ are the model parameters. The model's parameters are trained to minimize the negative log-likelihood over a dataset $D$ of input-out pairs, which in our case is the data of the four settings. Formally, we define the loss $\mathcal{L}_{\theta}$ as:
\begin{equation}
  \mathcal{L}_{\theta}(D) =  - \sum_j^{|D|} \sum_{i=0}^{n+m} \log p_{\theta}(x_i^{(j)}|x_{0}^{(j)}, \cdots, x_{i-1}^{(j)}), \label{eq:loss}
\end{equation}
where $n+m$ is a maximum sequence length in $D$. At inference time, given the input sequence $X$, the model parametrized by $\theta$ autoregressively generate the output sequence $Y$. 
\subsection{Continual Learning} \label{subsec:CL}
The aim of Continual Learning (CL) is to learn a set of tasks sequentially without catastrophically forgetting previously learned ones. In task-oriented dialogue systems, we cast CL as learning a sequence of domains sequentially. Let us define a curriculum of $T$ domains as the ordered set $\mathcal{D}=\{D_1, \cdots, D_T\}$, where $D_k$ is a dataset under the domain $K$. Moreover, we denote the models' parameter after learning the task $K$ with $\theta_K$. 

Following the recently defined taxonomy for CL~\cite{wortsman2020supermasks}, we study the settings in where the task-id is provided during training, but not during testing~\footnote{GNs: Task Given during train, Not inference; shared labels}. Meaning, during training the model is aware of which domain is currently learning, but during testing, the model is evaluated \textbf{without} specifying the dialogue domain. This assumption makes our CL setting more challenging but more realist, since during testing the users do not explicitly specify in which domain they want to operate. 

In this paper, we consider three Continual Learning approaches: \textit{regularization},
\textit{rehearsal} and \textit{architectural}. In our experiments, we describe the most commonly used methods for each approach, especially the one known to work well in language tasks. Thus:
\begin{itemize}[noitemsep]
    \item \textit{Regularization} methods add a regularization term to the current learned $\theta_t$ for avoiding to interfere with the previously learned once $\theta_{t-1}$. Formally the loss at task $t$ is:
    \begin{equation}
        L_{\theta_t}(D_t) = L_{\theta_t}(D_t) + \lambda \Omega (\theta_t - \theta^*_{t-1})^2
    \end{equation}
    where $\theta^*_{t-1}$ are a copy of the previously learned parameters frozen at this stage. In our experiments, we consider two kind of $\Omega$: the identity function (\textbf{L2}), and the fisher information matrix~\cite{kirkpatrick2017overcoming} (\textbf{EWC}).  
    
    \item \textit{Rehearsal} methods uses an episodic memory $\mathcal{M}$ to store examples from the previously learned domains, and re-use them while learning new tasks. The most straightforward methods is to add the content of the memory $\mathcal{M}$ to the current task data $D_t$. Following our notation, the model is optimized using $L_{\theta_t}(D_t+\mathcal{M})$, and we refer to this method as \textbf{REPLAY}. Another rehearsal method is to constrain the gradients updates so that the loss of the samples in memory never increases. More formally:
    \begin{equation}
        L_{\theta_t}(D_t) \ \text{s.t.} \ L_{\theta_t}(\mathcal{M}) \leq L_{\theta_{t-1}}(\mathcal{M}) 
    \end{equation}
    Of this kind, Gradient Episodic Memory (GEM)~\cite{lopez2017gradient} compute the gradient constrain via a quadratic programming solver that scales with the number of parameters of the model. After a first investigation, we discover that is impractical for large-language model to use GEM, since they have millions of parameters and the constrain is computed for each batch. To cope with this computational complexity, \citet{chaudhry2018efficient} proposed \textbf{A-GEM} that efficiently compute the gradient constrains while being effective in CL tasks. Finally, a rehearsal method specific to language is \textbf{LAMOL}~\cite{sun2019lamol} which instead of storing samples in $\mathcal{M}$, learn a model that simultaneously learns to solve the tasks and generate training samples.
    \item \textit{Architectural} methods add task-specific parameters to an existing base model for each task. Of this kind, multiple models has been proposed such as Progressive Net~\cite{rusu2016progressive}, Dynamically Expandable Networks (DEN)~\cite{yoon2017lifelong} and Learn-to-Grow~\cite{li2019learn}. On the other hand, there are fixed capacity methods, that do not add specific parameters but they learn parameters-masks~\cite{fernando2017pathnet}, usually binary~\cite{mallya2018piggyback}, to select sub-networks that are task-specific. To the best of our knowledge, these models have been tested mostly on computer vision tasks, and they can not easily handle our continual learning settings (i.e., no task-id during testing).
\end{itemize}

\begin{figure}[t]
    \centering
    \includegraphics[width=0.6\linewidth]{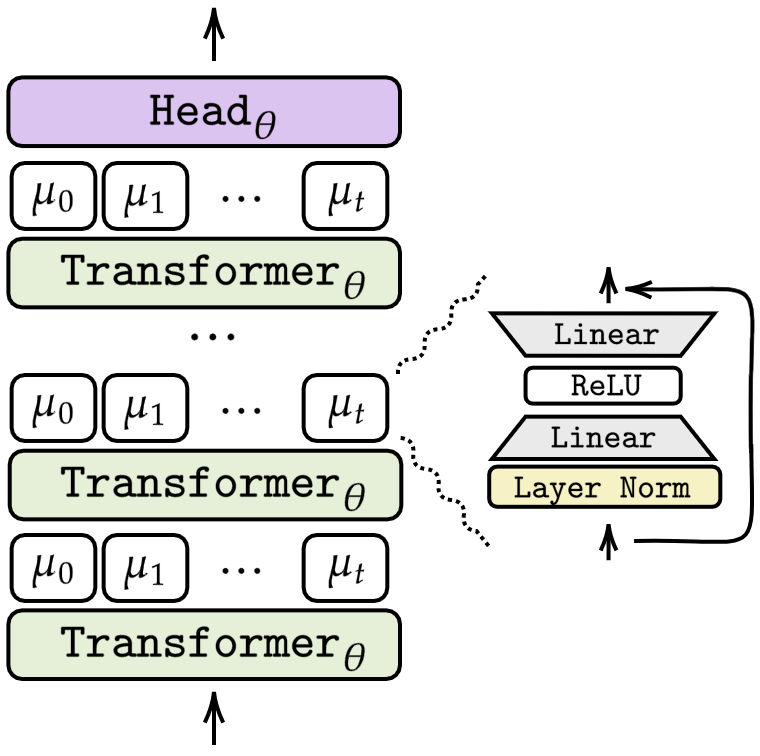}
    \caption{High-level representation of the AdapterCL. }
    \label{fig:adaptercl}
\end{figure}

\section{AdapterCL} \label{sec:adpt}
Motivated by the lack of architectural baselines for CL in sequence-to-sequence modelling, we propose a novel architectural method called AdapterCL. Our proposed method parameterizes each task using Residual Adapters~\cite{houlsby2019parameter} and uses an entropy-based classifier to select which adapter to use at testing time. This method is designed for large pre-trained language models, e.g. GPT-2, since only the task-specific parameter are trained while the original weights are left frozen. 

\paragraph{Residual adapters} are trainable parameters added on top of each transformer layer, which steer the output distribution of a pre-trained model without modifying its original weights. 
An adapter block consists of a Layer Normalization~\cite{ba2016layer}, followed by a two linear layer~\cite{hinton1994autoencoders} with a residual connection. Given the hidden representation at layer $l$ denoted as $H \in \mathbb{R}^{p\times d}$ of a transformer~\cite{vaswani2017attention}, where $d$ is the hidden size and $p$ is the sequence length, the residual adapter computes:
\begin{equation}
\mathrm{Adapter}_{\mu_i^l}(H) =  \mathrm{ReLU}(\mathrm{LN}(x) W_l^{E})W_l^{D} + H,
\end{equation}
where $W_l^{E}$ and $W_l^{D}$ are trainable parameters of dimensions $d\times b$ and $b\times d$ respectively, and \texttt{LN}$(\cdot)$ denotes the layer normalization. The bottleneck dimension $b$ is a tunable hyper-parameter that allows to adjust the capacity of the adapter according to the complexity of the target task. We define the set of $\mu_i = \{ W_0^{E}$, $W_0^{D}, \cdots,  W_L^{E}$, $W_L^{D} \}$ as the set of parameters for the Adapter$_i$ for a model with $L$ layers.

To continuously learning new tasks, we first spawn a new Adapter, parameterized by $\mu$, and then we train its parameters as in Equation~\ref{eq:loss}. For instance, given the dataset $D_t$ and the model with its corresponding Adapter $\mu_t$, the loss is defined as:
\begin{equation}
  \mathcal{L}_{\mu_t}(D_t) =  - \sum_j^{|D_t|} \sum_{i=0}^{n+m} \log p_{\mu_t}(x_i|x_{0}, \cdots x_{i-1}) \label{eq:lossadpt}
\end{equation}
Importantly, the loss is optimized over $\mu_t$ to guarantee that each task is independently learned. An high-level representation of AdapterCL is shown in Figure~\ref{fig:adaptercl}.

\paragraph{Perplexity-Based Classifier} In our CL setting the task-id is provided during training, thus each $\mu_t$ is optimized over $D_t$. During testing, instead, the task-id is not provided, thus the model has to predict which adapter to use for accomplishing the task. This step is not required in regularization and rehearsal approaches since a single set of parameters is optimised during training.

Inspired by \citet{wortsman2020supermasks}, we propose to utilize the perplexity of each adapter over the input $X$ as a measure of uncertainty. Thus, by selecting the adapter with lowest perplexity, we select the most confident model to generate the output sequence. The perplexity of an input sequence $X=x_{0}, \cdots, x_{n}$ is defined as:
\begin{equation}
    \mathrm{PPL}_{\theta}(X)=\sqrt[n]{\prod_{i=1}^{n} \frac{1}{p_{\theta}\left(x_{i} \mid x_{0}, \cdots, x_{i-1}\right)}}
\end{equation}
Therefore, given the set of adapters parametrized by $\mu_0,\dots, \mu_N$, each of which trained respectively with $D_0, \dots, D_N$, and an input sample $X$, we compute:
\begin{equation}
   \alpha_t = \mathrm{PPL}_{\mu_t}(X) \ \forall t \in 1, \cdots, N,\label{eq:alpha}
\end{equation}
in where each $\alpha_t$ represent the confidence of the adapter $t$ for the input $X$. The task-id $t$ is thus selected as 
\begin{equation}
    t^* = \operatorname*{argmin}{\alpha_0,\cdots,\alpha_N}\label{eq:selection}
\end{equation}
The perplexity-based selector requires a linear number of forwards with respect to the number of Adapters (Equation~\ref{eq:alpha}), but it has the advantage of not requiring a further classifier, which itself would suffer from catastrophic forgetting and would require an episodic memory. 

\section{Experimental Settings} \label{sec:experiment}
In this sections we describe 1) the datasets used for creating the learning curriculum, 2) the evaluation metric used to evaluate the different settings,and 3) the experimental setups.

\begin{table}[t]
\resizebox{\linewidth}{!}{
\begin{tabular}{r|ccc|ccc}
\hline
\textbf{Name}                          & \textbf{Train} & \textbf{Valid} & \textbf{Test} & \textbf{Dom.} & \textbf{Intent} & \textbf{Turns} \\ \hline
TM19 & 4,403           & 551            & 553           & 6             & 112           & 19.97          \\
TM20 & 13,839          & 1,731           & 1,734          & 7             & 128           & 16.92          \\
MWoZ & 7,906           & 1,000           & 1,000          & 5             & 15            & 13.93          \\
SGD & 5,278           & 761            & 1,531          & 19            & 43            & 14.71          \\ \hline
Total                         & 31,426          & 4,043           & 4,818          & 37            & 280           & 16.23          \\ \hline
\end{tabular}
}
\caption{Main datasets statistics. }
\label{tab:mainstat}
\end{table}

\subsection{Datasets}
To the best of our knowledge, there is no benchmark for CL in dialogue systems with a high number of tasks to be learned continuously and with multiple training settings. The closest to our is the work is from ~\citet{mi2020continual}, which continuously learn 5 domains in the NLG settings. In general, NLP benchmarks for continual learning uses no more that 10 tasks~\cite{sun2019lamol,d2019episodic}. Consequently, in this paper we propose a Continual Learning benchmark by jointly pre-processing four task-oriented dataset such as Task-Master 2019 (TM19)~\cite{byrne-etal-2019-taskmaster}, Task-Master 2020 (TM20)~\cite{byrne-etal-2019-taskmaster}, Schema Guided Dialogue (SGD)~\cite{rastogi2019towards} and MultiWoZ~\cite{budzianowski2018large}. This results in a curriculum of 37 domains to be learned continuously under four setting such as INTENT classification, DST, NLG, and E2E. This has been possible because the four datasets provides the speech-act annotation for both the user and the system turns, and the dialogue-state as well. For avoiding any domain overlapping, we select only the dialogues with a single domain and we do not merge domains with similar/same names or semantics. For example, the \textit{restaurant} domain appears in TM19, TM20 and MWOZ, with different slot names and values. Thus, we intentionally keep these data samples separate for modelling scenarios in which multiple apis are available for the same domain. 

Finally, the dataset are pre-processed as in Section~\ref{subsec:preproc} to form the four tasks, and the main statistics are shown in Table~\ref{tab:mainstat}. In appendix, Table~\ref{tab:all_data} we report the number of sample by domain and setting, where we can notice how different domains have a much different sample size, from few-hundred samples (e.g., SGD-travel) to 15K (e.g., TM19-flight). Importantly, since not all the datasets provides a delexicalized version of the responses, we decide to keep all the dataset in their plain text form.

\begin{table*}[t]
\centering
\begin{tabular}{r|cc|c|c|cc}
\hline
                                     & \multicolumn{1}{l}{} & \multicolumn{1}{l|}{} & \textbf{INTENT}              & \textbf{DST}            & \multicolumn{2}{c}{\textbf{NLG}}                     \\ \hline
\multicolumn{1}{c|}{\textbf{Method}} & \textbf{+Param.}     & \textbf{Mem.}         & \textit{Accuracy}$\uparrow$ & \textit{JGA}$\uparrow$ & \textit{EER}$\downarrow$ & \textit{BLEU}$\uparrow$ \\ \hline
{\ul \textit{VANILLA}}               & -                    &$\emptyset$          &4.08  $\pm$ 1.4              &4.91 $\pm$ 4.46         &48.73 $\pm$ 3.81           &6.38 $\pm$ 0.6          \\
{\ul \textit{L2}}                    &$|\theta|$          &$\emptyset$            &3.74  $\pm$ 1.4              &3.81 $\pm$ 3.44         &55.68 $\pm$ 7.09            &5.4 $\pm$ 0.9           \\
{\ul \textit{EWC}}                   &$2|\theta|$        &$\emptyset$             &3.95  $\pm$ 1.3              &5.22 $\pm$ 4.46         &58.2  $\pm$ 3.66           &5.06 $\pm$ 0.5          \\
{\ul \textit{AGEM}}                  & -                    &$t|M|$               &34.04 $\pm$ 6.36             &6.37 $\pm$ 4.0         &62.09 $\pm$ 6.88            &4.54 $\pm$ 0.6          \\
{\ul \textit{LAMOL}}                 & -                    &$\emptyset$          &7.49   $\pm$ 6.35            &4.55 $\pm$ 3.48        &66.11 $\pm$ 6.97             &3.0 $\pm$ 0.9           \\
{\ul \textit{REPLAY}}                & -                    &$t|M|$               &81.08 $\pm$ 1.37             &30.33$\pm$ 1.24 &\textbf{17.72} $\pm$ 0.85         &\textbf{17.4} $\pm$ 0.68         \\
{\ul \textit{ADAPT}}    &$t|\mu|$            &$\emptyset$          &      \textbf{90.46} $\pm$ 0.6    &\textbf{35.06} $\pm$ 0.52       &31.78 $\pm$ 1.28            &16.76 $\pm$ 0.34         \\ \hline
{\ul \textit{MULTI}}                 & -                    & -                     &95.45 $\pm$ 0.1            &48.9 $\pm$ 0.2        &12.56 $\pm$ 0.2             &23.61 $\pm$ 0.1         \\ \hline
\end{tabular}

\caption{E2E results in term of Intent accuracy, Joint-Goal-Accuracy (JGA), Slot-Error-Rate (EER) and BLUE. +Param shows the additional number of parameters per task, and Mem the episodic memory size need per task. }
\label{tab:e2e}
\end{table*}

\subsection{Evaluation Metric}
Automatic evaluations for end-to-end task-oriented dialogue systems is challenging, especially for the response generation. To overcome this issue, in this paper we use well-defined metric based on the three modularized settings. In each of the three sub-tasks, we define four metric as following:
\begin{itemize}[noitemsep]
    \item \textit{INTENT} recognition is evaluated using the \textit{accuracy} between the generated intent and the one provided in the gold label. 
    \item \textit{DST} is evaluated with the Joint Goal Accuracy (JGA)~\cite{wu2019transferable} over the gold dialogue state.
    \item \textit{NLG} is evaluated using both the BLEU score~\cite{papineni-etal-2002-bleu} and the slot error rate (EER)~\cite{wen2015semantically} which computed by ratio between the total number of slots and the values not appearing in the response. In datasets such as SGD, the slot has binary values e.g., yes or no, and thus we exclude those from the count as in ~\citet{kale2020few}. In the E2E setting, if the api output ($X_{out}$) is empty, then we rely on the BLEU score.  
\end{itemize}
Independently from the metric, we compute continual learning specific metrics such as the average metric thought time (Avg. Metric) as in ~\citet{lopez2017gradient}. We consider access to the test set for each of the $T$ tasks and after the model finishes learning the task $t_i$, we evaluate the model test performance on all tasks in the curriculum. By doing so, we construct the matrix $R\in \mathbb{R}^{T\times T}$, where $R_{i,j}$ is the test metric (e.g., BLEU, JGA) of the model on task $t_j$ after observing the last sample from task $t_i$. Then we define the average accuracy as:
\begin{equation}
    \mathrm{Avg. Metric} =\frac{1}{T} \sum_{i=1}^{T} R_{T, i} \label{eq:avgmetric}
\end{equation}
The Avg. Metric score is useful for understanding the learning dynamics thought time of different baselines. Further metrics such as Backward-Transfer and Forward-Transfer~\cite{lopez2017gradient} are available to distinguish baselines with similar Avg. Metric score, but in this paper we limit our evaluation to this metric, since there is a large gap among different baselines. Finally, to evaluate the adapter selection, we use the accuracy over the gold task-id.

\subsection{Baselines and Settings}
The main aim of this paper is to compare the performance of different CL approaches and to understand what are the trade-off among them. Therefore, following the definition provided in Section~\ref{subsec:CL}, we compare: 1) EWC and L2, 2) A-GEM, LAMOL, and REPLAY, and 3) AdapterCL. Additionally, we provide a baselines trained on the task continuously, namely VANILLA, without any regularization or memory, and a multitask baseline (MULTI), which is trained on all the data in the curriculum at the same time. In L2, EWC, and A-GEM we tune different $\lambda$ in the range 0.0001 to 100, and in \textit{rehearsal} based methods, such as REPLAY and GEM, we keep 50 samples per task, for a total of 1850 sample in $\mathcal{M}$ at the end of the curriculum. This is particularly important since if we store in memory all the samples of the seen tasks the model require an high training cost. Arguably, this could be an option if the per-task sample size is small, but this is not always possible as in large language models~\cite{brown2020language}. Therefore the assumption of minimizing the number of samples in memory is valid and widely used in the CL literature~\cite{mi2020continual}. Finally, for the AdapterCL, we tune the bottleneck size $b$ between 10, 50, 100, 200. Interested readers can refer to appendix for further details.   
In Continual Learning the model is not able decide the order of tasks, therefore, we create 5 learning curriculum by randomly permuting the 37 tasks.

\begin{figure}[t]
    \centering
    \includegraphics[width=\linewidth]{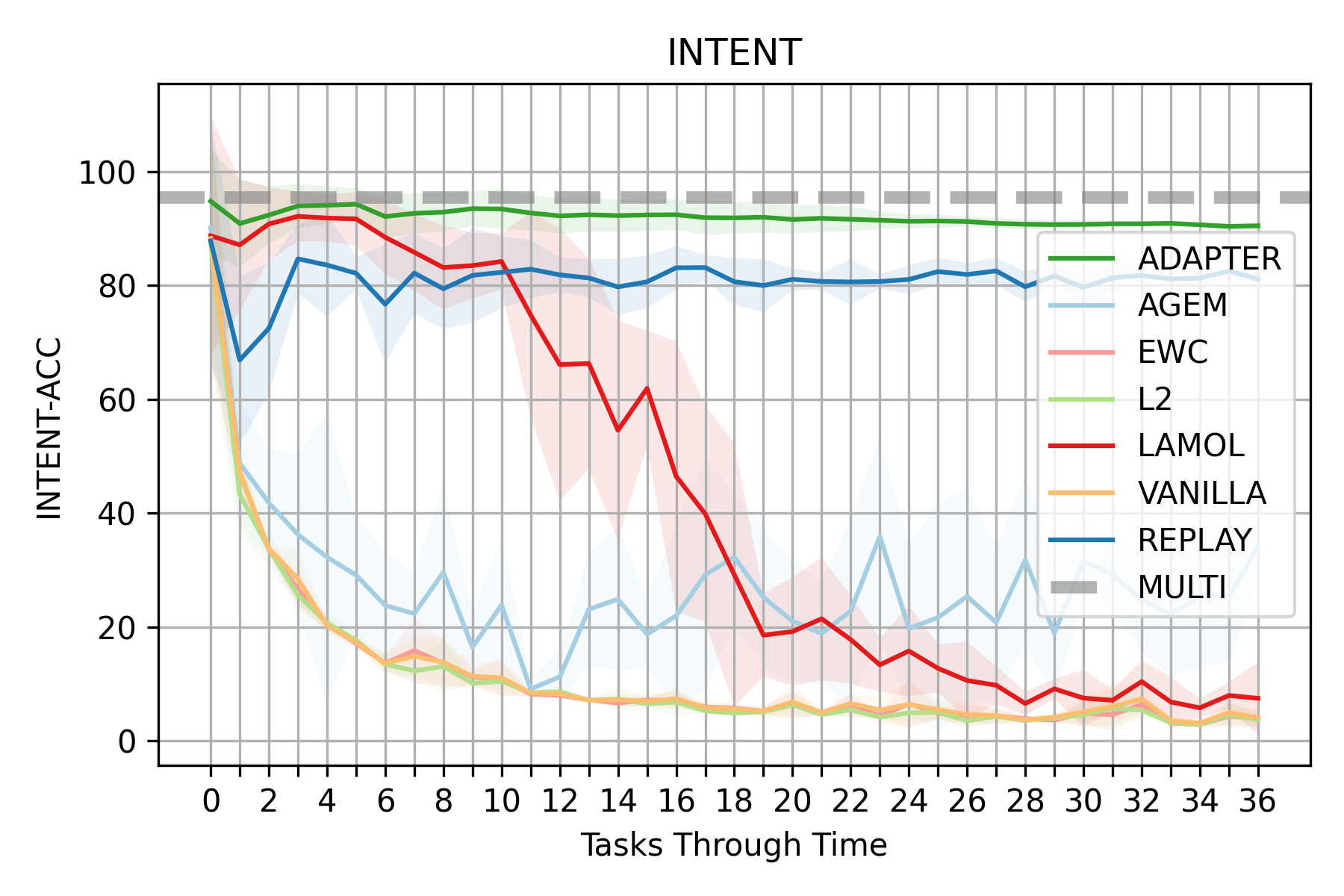}
    \includegraphics[width=\linewidth]{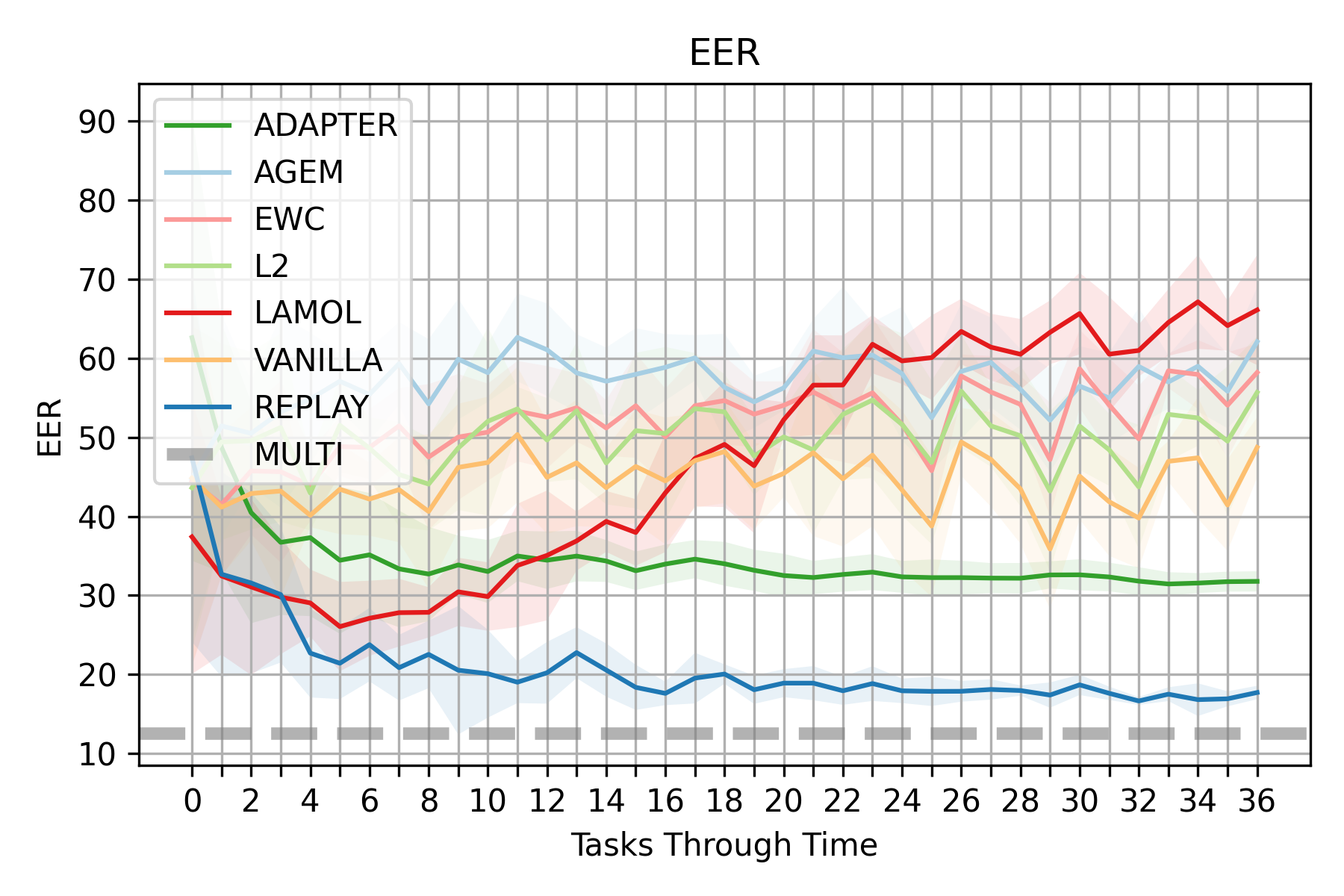}
    \caption{Avg. Metric for the Intent Accuracy and EER in the E2E setting. }
    \label{fig:E2E}
\end{figure}

\section{Results \& Analysis}\label{sec:results}
The main results in the end-to-end setting are summarized in Table~\ref{tab:e2e}, while the one from the modularized settings in Table~\ref{tab:modularized_results} in the Appendix. In these tables we report the Avg. Metric at the end of the curriculum, which is equivalent to the average test set performance in all the tasks, and the resources used by each model. 

\paragraph{Main Results} From these tables, we can observe that: 1) both regularization-based methods (L2/EWC) and some rehearsal-based methods (AGEM/LAMOL) suffers greatly in this task, 2) REPLAY and AdapterCL perform comparably well in the Intent and DST metric, 3) REPLAY works the best in the NLG task, showing that transferring learning between tasks is needed, and 4) no CL methods can reach the multi-task baseline, especially in the DST task. Moreover, the adapter selection accuracy based on Equation~\ref{eq:selection}, is of 95.44$\pm$0.2\% in end-to-end, 98.03$\pm$0.1\% in Intent Recognition, 98.19$\pm$0.1\% in DST, and 93.98$\pm$0.1\%.

Although these numbers are meaningful, they do not describe the entire learning history of the curriculum. To better understand this dynamics, we plot the Avg. Metric in Equation~\ref{eq:avgmetric} after each task is learned (i.e. t=T in the equation). Figure~\ref{fig:E2E} shows the plot for the two of the considered metrics and all the baselines. From this figure we can better understand how REPLAY and AdapterCL outperforms other baselines and, interestingly, that LAMOL performs as well as replay in the first 12 tasks. This because LAMOL learns to generate training samples instead of using an explicit memory, and thus the generation become harder when more and more task are shown. This further strength our motivation for having a benchmark with a long curriculum. In Appendix, Figure~\ref{fig:E2E2} shows the remain two metrics for the E2E setting and Figure~\ref{fig:Module} show the same plots for the individual modules training. 

\subsection{No Free Lunch} Finally based on the results shown in Table~\ref{tab:e2e}, especially based on the resource used by each methods, we conclude that there is a no free lunch in term of resources needed for avoiding the catastrophic problem. To elaborate, in both REPLAY and AdapterCL the resource used grows linearly with the number of tasks, i.e., in REPLAY the number of samples stored in the episodic memory grows linearly (i.e., 50 times the number of tasks), and in AdapterCL the number of parameters grows linearly (i.e., number of adapter parameter times the number of tasks). Figure~\ref{fig:freelunch} in Appendix describe an high-level intuition of this concept by plotting the number of tasks vs parameter and episodic memory size.

Therefore, given a resource budget, in term of memory or parameters, different baselines are preferable. The main advantage of using memory-base (e.g., REPLAY) that no parameters are added thus the resulting model is closer to a multitask baseline, but it comes with the disadvantage of loosing the weights of the original pre-trained model. This is particularly critical for large pre-trained LMs which provide a good starting points for fine-tuning new tasks. On the other hand, the main advantage of parameters isolation methods (e.g., AdapterCL) is the ability to retaining the original weights and the ability of controlling which task to trigger given a certain input. The latter is important in scenarios where just a subset of domains are shown to the user (e.g. only one particular restaurant api). The main disadvantage, instead, is the lack of knowledge transfer among tasks, since each dataset is trained in isolation. 

\begin{figure}[t]
    \centering
    \includegraphics[width=\linewidth]{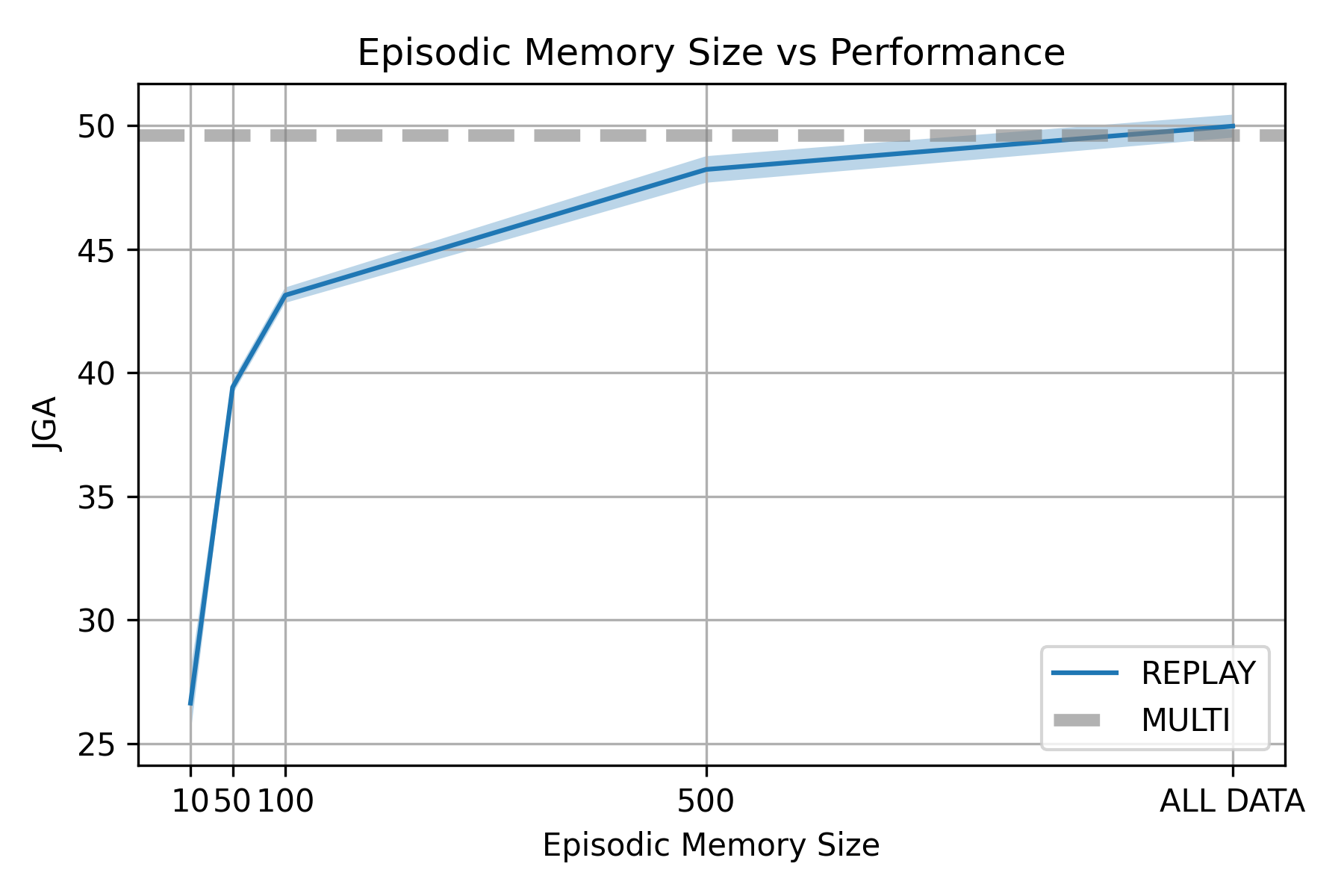}
    \caption{Ablation study on the size of the episodic-memory vs JGA. }
    \label{fig:JGAMEM}
\end{figure}

\subsection{Analysis: Episodic Memory Size}
In this section, we analyses the effect of increasing the episodic memory size for the REPLAY method. Trivially, by including all the training sample in the memory, the model at the last task converges to the multitask baseline. Then, the question of many how sample to keep per task for avoiding catastrophic forgetting is important. In light of this, Figure~\ref{fig:JGAMEM} shows the performance of different Episodic Memory size in the DST task. Here we observe that by storing only few samples per task (i.e., 10-50) the model still greatly suffers from catastrophic forgetting, where instead with around 500 samples, which is equivalent to a total of 18,500 sample in our setting, the performance are closer to the multitask baseline (i.e., a possible upper bound). Similar observation are shown for the other two tasks Figure~\ref{fig:intentmem} ~\ref{fig:BLUEEER} in Appendix. 

\section{Related Work}
Continual Learning methods are usually developed and benchmark on computer visions tasks. Interested readers may refer to ~\citet{mundt2020wholistic,parisi2019continual,de2019continual} for an overview of existing approaches, and to Section~\ref{subsec:CL} for more details on the three main CL approaches studied in this paper. On the other hand, Continual Learning is also studied in the Long Life Learning (LLL) scenario, where a learner continuously accumulate knowledge and makes use of it in the future~\cite{chen2018lifelong,liu2020lifelong}. In this paper, we study the setting in where a series of task are learned continuously.

\paragraph{Continual Learning in NLP} has been explore in both in classification ~\cite{d2019episodic,sprechmann2018memory,wang2020efficient} and generation~\cite{sun2019lamol,hu2020drinking} tasks. For instance, \citet{sun2019lamol,chuang2020lifelong} proposed LAMOL, which we use as our baselines, and studied its effectiveness in a subset of DecaNLP~\cite{mccann2018natural}. On the other hand, the work of ~\citet{d2019episodic,sprechmann2018memory} are not suitable for interactive systems as dialogue systems, since they require a local adaptation (i.e., a fine-tuning step) during inference. Finally, continual learning has been used for sentence encoder~\cite{liu2019continual}, composition language learning~\cite{li2019compositional} and relation learning ~\cite{han2020continual}. However, these methods are very specific to particular NLP applications. 

\paragraph{Continual Learning in Dialogue Systems} The very early work on CL for Task-Oriented dialogue is from ~\citet{lee2017toward}, where EWC is used to avoid catastrophic forgetting on three domains learned sequentially. Continual Learning has been studied in the NLG setting, where a single model is trained to learn one domain at the time in MWoZ~\cite{mi2020continual}. The author, used a episodic memory to replay the example in combination with EWC. In this paper, we compare similar baselines, in a large benchmark that also include MWoZ and the NLG settings. Continual Learning has been studied also in the DST setting in ~\cite{wu2019transferable} using MWoZ, where several baselines such as L2, EWC and GEM has been compared. Differently, \cite{li2019evaluate} leverage CL for evaluate the quality of chat-bot models and \citet{he2019mix} studied the catastrophic forgetting problem in Chit-Chat systems. Finally, \cite{shuster2020deploying} showed that by training models on humans-machine conversations in a open-domain fantasy world game~\cite{fan2020generating} the models progressively improve, as measured by automatic metrics and online engagement scores.

\section{Conclusion}
In this paper, we proposed a benchmark for Continual Learning in task-oriented dialogue systems, with 37 tasks to be learned continuously on four settings such as Intent recognition, Dialogue State Tracking, Natural Language Generation, and end-to-end. Then, we implemented three different Continual Learning methodologies such as regularization, rehearsal and architectural. In the latter, we propose a simple yet effective methods based on residual adapters and uses an entropy-based classifier to select which adapter to use at testing time. Finally, we analyse the trade-off between performance, number-of-parameters, and episodic memories size of the evaluated baselines, unveiling a no-free lunch among this methods.

\bibliography{anthology,acl2020}

\begin{thebibliography}{49}
\expandafter\ifx\csname natexlab\endcsname\relax\def\natexlab#1{#1}\fi

\bibitem[{Ba et~al.(2016)Ba, Kiros, and Hinton}]{ba2016layer}
Jimmy~Lei Ba, Jamie~Ryan Kiros, and Geoffrey~E Hinton. 2016.
\newblock Layer normalization.
\newblock \emph{arXiv preprint arXiv:1607.06450}.

\bibitem[{Bengio et~al.(2003)Bengio, Ducharme, Vincent, and
  Jauvin}]{bengio2003neural}
Yoshua Bengio, R{\'e}jean Ducharme, Pascal Vincent, and Christian Jauvin. 2003.
\newblock A neural probabilistic language model.
\newblock \emph{Journal of machine learning research}, 3(Feb):1137--1155.

\bibitem[{Brown et~al.(2020)Brown, Mann, Ryder, Subbiah, Kaplan, Dhariwal,
  Neelakantan, Shyam, Sastry, Askell et~al.}]{brown2020language}
Tom~B Brown, Benjamin Mann, Nick Ryder, Melanie Subbiah, Jared Kaplan, Prafulla
  Dhariwal, Arvind Neelakantan, Pranav Shyam, Girish Sastry, Amanda Askell,
  et~al. 2020.
\newblock Language models are few-shot learners.
\newblock \emph{arXiv preprint arXiv:2005.14165}.

\bibitem[{Budzianowski et~al.(2018)Budzianowski, Wen, Tseng, Casanueva, Stefan,
  Osman, and Ga{\v{s}}i\'c}]{budzianowski2018large}
Pawe{\l} Budzianowski, Tsung-Hsien Wen, Bo-Hsiang Tseng, I{\~n}igo Casanueva,
  Ultes Stefan, Ramadan Osman, and Milica Ga{\v{s}}i\'c. 2018.
\newblock Multiwoz - a large-scale multi-domain wizard-of-oz dataset for
  task-oriented dialogue modelling.
\newblock In \emph{Proceedings of the 2018 Conference on Empirical Methods in
  Natural Language Processing (EMNLP)}.

\bibitem[{Byrne et~al.(2020)Byrne, Krishnamoorthi, Ganesh, and
  Kale}]{byrne2020tickettalk}
Bill Byrne, Karthik Krishnamoorthi, Saravanan Ganesh, and Mihir~Sanjay Kale.
  2020.
\newblock Tickettalk: Toward human-level performance with end-to-end,
  transaction-based dialog systems.
\newblock \emph{arXiv preprint arXiv:2012.12458}.

\bibitem[{Byrne et~al.(2019)Byrne, Krishnamoorthi, Sankar, Neelakantan,
  Duckworth, Yavuz, Goodrich, Dubey, Kim, and
  Cedilnik}]{byrne-etal-2019-taskmaster}
Bill Byrne, Karthik Krishnamoorthi, Chinnadhurai Sankar, Arvind Neelakantan,
  Daniel Duckworth, Semih Yavuz, Ben Goodrich, Amit Dubey, Kyu-Young Kim, and
  Andy Cedilnik. 2019.
\newblock Taskmaster-1:toward a realistic and diverse dialog dataset.
\newblock In \emph{2019 Conference on Empirical Methods in Natural Language
  Processing and 9th International Joint Conference on Natural Language
  Processing}, Hong Kong.

\bibitem[{Chaudhry et~al.(2018)Chaudhry, Ranzato, Rohrbach, and
  Elhoseiny}]{chaudhry2018efficient}
Arslan Chaudhry, Marc'Aurelio Ranzato, Marcus Rohrbach, and Mohamed Elhoseiny.
  2018.
\newblock Efficient lifelong learning with a-gem.
\newblock \emph{arXiv preprint arXiv:1812.00420}.

\bibitem[{Chen and Liu(2018)}]{chen2018lifelong}
Zhiyuan Chen and Bing Liu. 2018.
\newblock Lifelong machine learning.
\newblock \emph{Synthesis Lectures on Artificial Intelligence and Machine
  Learning}, 12(3):1--207.

\bibitem[{Chuang et~al.(2020)Chuang, Su, and Chen}]{chuang2020lifelong}
Yung-Sung Chuang, Shang-Yu Su, and Yun-Nung Chen. 2020.
\newblock Lifelong language knowledge distillation.
\newblock \emph{arXiv preprint arXiv:2010.02123}.

\bibitem[{d'Autume et~al.(2019)d'Autume, Ruder, Kong, and
  Yogatama}]{d2019episodic}
Cyprien de~Masson d'Autume, Sebastian Ruder, Lingpeng Kong, and Dani Yogatama.
  2019.
\newblock Episodic memory in lifelong language learning.
\newblock \emph{arXiv preprint arXiv:1906.01076}.

\bibitem[{De~Lange et~al.(2019)De~Lange, Aljundi, Masana, Parisot, Jia,
  Leonardis, Slabaugh, and Tuytelaars}]{de2019continual}
Matthias De~Lange, Rahaf Aljundi, Marc Masana, Sarah Parisot, Xu~Jia,
  Ale{\v{s}} Leonardis, Gregory Slabaugh, and Tinne Tuytelaars. 2019.
\newblock A continual learning survey: Defying forgetting in classification
  tasks.
\newblock \emph{arXiv preprint arXiv:1909.08383}.

\bibitem[{Fan et~al.(2020)Fan, Urbanek, Ringshia, Dinan, Qian, Karamcheti,
  Prabhumoye, Kiela, Rockt{\"a}schel, Szlam et~al.}]{fan2020generating}
Angela Fan, Jack Urbanek, Pratik Ringshia, Emily Dinan, Emma Qian, Siddharth
  Karamcheti, Shrimai Prabhumoye, Douwe Kiela, Tim Rockt{\"a}schel, Arthur
  Szlam, et~al. 2020.
\newblock Generating interactive worlds with text.
\newblock In \emph{AAAI}, pages 1693--1700.

\bibitem[{Fernando et~al.(2017)Fernando, Banarse, Blundell, Zwols, Ha, Rusu,
  Pritzel, and Wierstra}]{fernando2017pathnet}
Chrisantha Fernando, Dylan Banarse, Charles Blundell, Yori Zwols, David Ha,
  Andrei~A Rusu, Alexander Pritzel, and Daan Wierstra. 2017.
\newblock Pathnet: Evolution channels gradient descent in super neural
  networks.
\newblock \emph{arXiv preprint arXiv:1701.08734}.

\bibitem[{Han et~al.(2020)Han, Dai, Gao, Lin, Liu, Li, Sun, and
  Zhou}]{han2020continual}
Xu~Han, Yi~Dai, Tianyu Gao, Yankai Lin, Zhiyuan Liu, Peng Li, Maosong Sun, and
  Jie Zhou. 2020.
\newblock Continual relation learning via episodic memory activation and
  reconsolidation.
\newblock In \emph{Proceedings of the 58th Annual Meeting of the Association
  for Computational Linguistics}, pages 6429--6440.

\bibitem[{He et~al.(2019)He, Liu, Cho, Ott, Liu, Glass, and Peng}]{he2019mix}
Tianxing He, Jun Liu, Kyunghyun Cho, Myle Ott, Bing Liu, James Glass, and
  Fuchun Peng. 2019.
\newblock Mix-review: Alleviate forgetting in the pretrain-finetune framework
  for neural language generation models.
\newblock \emph{arXiv preprint arXiv:1910.07117}.

\bibitem[{Hinton and Zemel(1994)}]{hinton1994autoencoders}
Geoffrey~E Hinton and Richard~S Zemel. 1994.
\newblock Autoencoders, minimum description length and helmholtz free energy.
\newblock In \emph{Advances in neural information processing systems}, pages
  3--10.

\bibitem[{Hosseini-Asl et~al.(2020)Hosseini-Asl, McCann, Wu, Yavuz, and
  Socher}]{hosseini2020simple}
Ehsan Hosseini-Asl, Bryan McCann, Chien-Sheng Wu, Semih Yavuz, and Richard
  Socher. 2020.
\newblock A simple language model for task-oriented dialogue.
\newblock \emph{arXiv preprint arXiv:2005.00796}.

\bibitem[{Houlsby et~al.(2019)Houlsby, Giurgiu, Jastrzebski, Morrone,
  De~Laroussilhe, Gesmundo, Attariyan, and Gelly}]{houlsby2019parameter}
Neil Houlsby, Andrei Giurgiu, Stanislaw Jastrzebski, Bruna Morrone, Quentin
  De~Laroussilhe, Andrea Gesmundo, Mona Attariyan, and Sylvain Gelly. 2019.
\newblock Parameter-efficient transfer learning for nlp.
\newblock \emph{arXiv preprint arXiv:1902.00751}.

\bibitem[{Hu et~al.(2020)Hu, Sener, Sha, and Koltun}]{hu2020drinking}
Hexiang Hu, Ozan Sener, Fei Sha, and Vladlen Koltun. 2020.
\newblock Drinking from a firehose: Continual learning with web-scale natural
  language.
\newblock \emph{arXiv preprint arXiv:2007.09335}.

\bibitem[{Kale and Rastogi(2020)}]{kale2020few}
Mihir Kale and Abhinav Rastogi. 2020.
\newblock Few-shot natural language generation by rewriting templates.
\newblock \emph{arXiv preprint arXiv:2004.15006}.

\bibitem[{Kirkpatrick et~al.(2017)Kirkpatrick, Pascanu, Rabinowitz, Veness,
  Desjardins, Rusu, Milan, Quan, Ramalho, Grabska-Barwinska
  et~al.}]{kirkpatrick2017overcoming}
James Kirkpatrick, Razvan Pascanu, Neil Rabinowitz, Joel Veness, Guillaume
  Desjardins, Andrei~A Rusu, Kieran Milan, John Quan, Tiago Ramalho, Agnieszka
  Grabska-Barwinska, et~al. 2017.
\newblock Overcoming catastrophic forgetting in neural networks.
\newblock \emph{Proceedings of the national academy of sciences},
  114(13):3521--3526.

\bibitem[{Lee(2017)}]{lee2017toward}
Sungjin Lee. 2017.
\newblock Toward continual learning for conversational agents.
\newblock \emph{arXiv preprint arXiv:1712.09943}.

\bibitem[{Lei et~al.(2018)Lei, Jin, Kan, Ren, He, and Yin}]{lei2018sequicity}
Wenqiang Lei, Xisen Jin, Min-Yen Kan, Zhaochun Ren, Xiangnan He, and Dawei Yin.
  2018.
\newblock Sequicity: Simplifying task-oriented dialogue systems with single
  sequence-to-sequence architectures.
\newblock In \emph{Proceedings of the 56th Annual Meeting of the Association
  for Computational Linguistics (Volume 1: Long Papers)}, pages 1437--1447.

\bibitem[{Li et~al.(2019{\natexlab{a}})Li, He, Zhou, and Yu}]{li2019evaluate}
Lu~Li, Zhongheng He, Xiangyang Zhou, and Dianhai Yu. 2019{\natexlab{a}}.
\newblock How to evaluate the next system: Automatic dialogue evaluation from
  the perspective of continual learning.
\newblock \emph{arXiv preprint arXiv:1912.04664}.

\bibitem[{Li et~al.(2019{\natexlab{b}})Li, Zhou, Wu, Socher, and
  Xiong}]{li2019learn}
Xilai Li, Yingbo Zhou, Tianfu Wu, Richard Socher, and Caiming Xiong.
  2019{\natexlab{b}}.
\newblock Learn to grow: A continual structure learning framework for
  overcoming catastrophic forgetting.
\newblock \emph{arXiv preprint arXiv:1904.00310}.

\bibitem[{Li et~al.(2019{\natexlab{c}})Li, Zhao, Church, and
  Elhoseiny}]{li2019compositional}
Yuanpeng Li, Liang Zhao, Kenneth Church, and Mohamed Elhoseiny.
  2019{\natexlab{c}}.
\newblock Compositional language continual learning.
\newblock In \emph{International Conference on Learning Representations}.

\bibitem[{Liu and Mei(2020)}]{liu2020lifelong}
Bing Liu and Chuhe Mei. 2020.
\newblock Lifelong knowledge learning in rule-based dialogue systems.
\newblock \emph{arXiv preprint arXiv:2011.09811}.

\bibitem[{Liu et~al.(2019)Liu, Ungar, and Sedoc}]{liu2019continual}
Tianlin Liu, Lyle Ungar, and Jo{\~a}o Sedoc. 2019.
\newblock Continual learning for sentence representations using conceptors.
\newblock \emph{arXiv preprint arXiv:1904.09187}.

\bibitem[{Lopez-Paz and Ranzato(2017)}]{lopez2017gradient}
David Lopez-Paz and Marc'Aurelio Ranzato. 2017.
\newblock Gradient episodic memory for continual learning.
\newblock In \emph{Advances in neural information processing systems}, pages
  6467--6476.

\bibitem[{Mallya et~al.(2018)Mallya, Davis, and Lazebnik}]{mallya2018piggyback}
Arun Mallya, Dillon Davis, and Svetlana Lazebnik. 2018.
\newblock Piggyback: Adapting a single network to multiple tasks by learning to
  mask weights.
\newblock In \emph{Proceedings of the European Conference on Computer Vision
  (ECCV)}, pages 67--82.

\bibitem[{McCann et~al.(2018)McCann, Keskar, Xiong, and
  Socher}]{mccann2018natural}
Bryan McCann, Nitish~Shirish Keskar, Caiming Xiong, and Richard Socher. 2018.
\newblock The natural language decathlon: Multitask learning as question
  answering.
\newblock \emph{arXiv preprint arXiv:1806.08730}.

\bibitem[{McCloskey and Cohen(1989)}]{mccloskey1989catastrophic}
Michael McCloskey and Neal~J Cohen. 1989.
\newblock Catastrophic interference in connectionist networks: The sequential
  learning problem.
\newblock In \emph{Psychology of learning and motivation}, volume~24, pages
  109--165. Elsevier.

\bibitem[{Mi et~al.(2020)Mi, Chen, Zhao, Huang, and Faltings}]{mi2020continual}
Fei Mi, Liangwei Chen, Mengjie Zhao, Minlie Huang, and Boi Faltings. 2020.
\newblock Continual learning for natural language generation in task-oriented
  dialog systems.
\newblock \emph{arXiv preprint arXiv:2010.00910}.

\bibitem[{Mundt et~al.(2020)Mundt, Hong, Pliushch, and
  Ramesh}]{mundt2020wholistic}
Martin Mundt, Yong~Won Hong, Iuliia Pliushch, and Visvanathan Ramesh. 2020.
\newblock A wholistic view of continual learning with deep neural networks:
  Forgotten lessons and the bridge to active and open world learning.
\newblock \emph{arXiv preprint arXiv:2009.01797}.

\bibitem[{Papineni et~al.(2002)Papineni, Roukos, Ward, and
  Zhu}]{papineni-etal-2002-bleu}
Kishore Papineni, Salim Roukos, Todd Ward, and Wei-Jing Zhu. 2002.
\newblock \href {https://doi.org/10.3115/1073083.1073135} {{B}leu: a method for
  automatic evaluation of machine translation}.
\newblock In \emph{Proceedings of the 40th Annual Meeting of the Association
  for Computational Linguistics}, pages 311--318, Philadelphia, Pennsylvania,
  USA. Association for Computational Linguistics.

\bibitem[{Parisi et~al.(2019)Parisi, Kemker, Part, Kanan, and
  Wermter}]{parisi2019continual}
German~I Parisi, Ronald Kemker, Jose~L Part, Christopher Kanan, and Stefan
  Wermter. 2019.
\newblock Continual lifelong learning with neural networks: A review.
\newblock \emph{Neural Networks}, 113:54--71.

\bibitem[{Peng et~al.(2020)Peng, Li, Li, Shayandeh, Liden, and
  Gao}]{peng2020soloist}
Baolin Peng, Chunyuan Li, Jinchao Li, Shahin Shayandeh, Lars Liden, and
  Jianfeng Gao. 2020.
\newblock Soloist: Few-shot task-oriented dialog with a single pre-trained
  auto-regressive model.
\newblock \emph{arXiv preprint arXiv:2005.05298}.

\bibitem[{Rastogi et~al.(2019)Rastogi, Zang, Sunkara, Gupta, and
  Khaitan}]{rastogi2019towards}
Abhinav Rastogi, Xiaoxue Zang, Srinivas Sunkara, Raghav Gupta, and Pranav
  Khaitan. 2019.
\newblock Towards scalable multi-domain conversational agents: The
  schema-guided dialogue dataset.
\newblock \emph{arXiv preprint arXiv:1909.05855}.

\bibitem[{Rusu et~al.(2016)Rusu, Rabinowitz, Desjardins, Soyer, Kirkpatrick,
  Kavukcuoglu, Pascanu, and Hadsell}]{rusu2016progressive}
Andrei~A Rusu, Neil~C Rabinowitz, Guillaume Desjardins, Hubert Soyer, James
  Kirkpatrick, Koray Kavukcuoglu, Razvan Pascanu, and Raia Hadsell. 2016.
\newblock Progressive neural networks.
\newblock \emph{arXiv preprint arXiv:1606.04671}.

\bibitem[{Shuster et~al.(2020)Shuster, Urbanek, Dinan, Szlam, and
  Weston}]{shuster2020deploying}
Kurt Shuster, Jack Urbanek, Emily Dinan, Arthur Szlam, and Jason Weston. 2020.
\newblock Deploying lifelong open-domain dialogue learning.
\newblock \emph{arXiv preprint arXiv:2008.08076}.

\bibitem[{Sprechmann et~al.(2018)Sprechmann, Jayakumar, Rae, Pritzel, Badia,
  Uria, Vinyals, Hassabis, Pascanu, and Blundell}]{sprechmann2018memory}
Pablo Sprechmann, Siddhant~M Jayakumar, Jack~W Rae, Alexander Pritzel,
  Adria~Puigdomenech Badia, Benigno Uria, Oriol Vinyals, Demis Hassabis, Razvan
  Pascanu, and Charles Blundell. 2018.
\newblock Memory-based parameter adaptation.
\newblock \emph{arXiv preprint arXiv:1802.10542}.

\bibitem[{Sun et~al.(2019)Sun, Ho, and Lee}]{sun2019lamol}
Fan-Keng Sun, Cheng-Hao Ho, and Hung-Yi Lee. 2019.
\newblock Lamol: Language modeling for lifelong language learning.
\newblock In \emph{International Conference on Learning Representations}.

\bibitem[{Thrun and Pratt(2012)}]{thrun2012learning}
Sebastian Thrun and Lorien Pratt. 2012.
\newblock \emph{Learning to learn}.
\newblock Springer Science \& Business Media.

\bibitem[{Vaswani et~al.(2017)Vaswani, Shazeer, Parmar, Uszkoreit, Jones,
  Gomez, Kaiser, and Polosukhin}]{vaswani2017attention}
Ashish Vaswani, Noam Shazeer, Niki Parmar, Jakob Uszkoreit, Llion Jones,
  Aidan~N Gomez, {\L}ukasz Kaiser, and Illia Polosukhin. 2017.
\newblock Attention is all you need.
\newblock In \emph{Advances in neural information processing systems}, pages
  5998--6008.

\bibitem[{Wang et~al.(2020)Wang, Mehta, P{\'o}czos, and
  Carbonell}]{wang2020efficient}
Zirui Wang, Sanket~Vaibhav Mehta, Barnab{\'a}s P{\'o}czos, and Jaime Carbonell.
  2020.
\newblock Efficient meta lifelong-learning with limited memory.
\newblock \emph{arXiv preprint arXiv:2010.02500}.

\bibitem[{Wen et~al.(2015)Wen, Gasic, Mrksic, Su, Vandyke, and
  Young}]{wen2015semantically}
Tsung-Hsien Wen, Milica Gasic, Nikola Mrksic, Pei-Hao Su, David Vandyke, and
  Steve Young. 2015.
\newblock Semantically conditioned lstm-based natural language generation for
  spoken dialogue systems.
\newblock \emph{arXiv preprint arXiv:1508.01745}.

\bibitem[{Wortsman et~al.(2020)Wortsman, Ramanujan, Liu, Kembhavi, Rastegari,
  Yosinski, and Farhadi}]{wortsman2020supermasks}
Mitchell Wortsman, Vivek Ramanujan, Rosanne Liu, Aniruddha Kembhavi, Mohammad
  Rastegari, Jason Yosinski, and Ali Farhadi. 2020.
\newblock Supermasks in superposition.
\newblock \emph{Advances in Neural Information Processing Systems}, 33.

\bibitem[{Wu et~al.(2019)Wu, Madotto, Hosseini-Asl, Xiong, Socher, and
  Fung}]{wu2019transferable}
Chien-Sheng Wu, Andrea Madotto, Ehsan Hosseini-Asl, Caiming Xiong, Richard
  Socher, and Pascale Fung. 2019.
\newblock Transferable multi-domain state generator for task-oriented dialogue
  systems.
\newblock \emph{arXiv preprint arXiv:1905.08743}.

\bibitem[{Yoon et~al.(2017)Yoon, Yang, Lee, and Hwang}]{yoon2017lifelong}
Jaehong Yoon, Eunho Yang, Jeongtae Lee, and Sung~Ju Hwang. 2017.
\newblock Lifelong learning with dynamically expandable networks.
\newblock \emph{arXiv preprint arXiv:1708.01547}.

\end{thebibliography}
\bibliographystyle{acl_natbib}

\appendix

\section{Appendices}
\label{sec:appendix}

\section{Supplemental Material}
\label{sec:supplemental}
\begin{figure}[t]
    \centering
    \includegraphics[width=\linewidth]{E2E_ACC.png}
    \includegraphics[width=\linewidth]{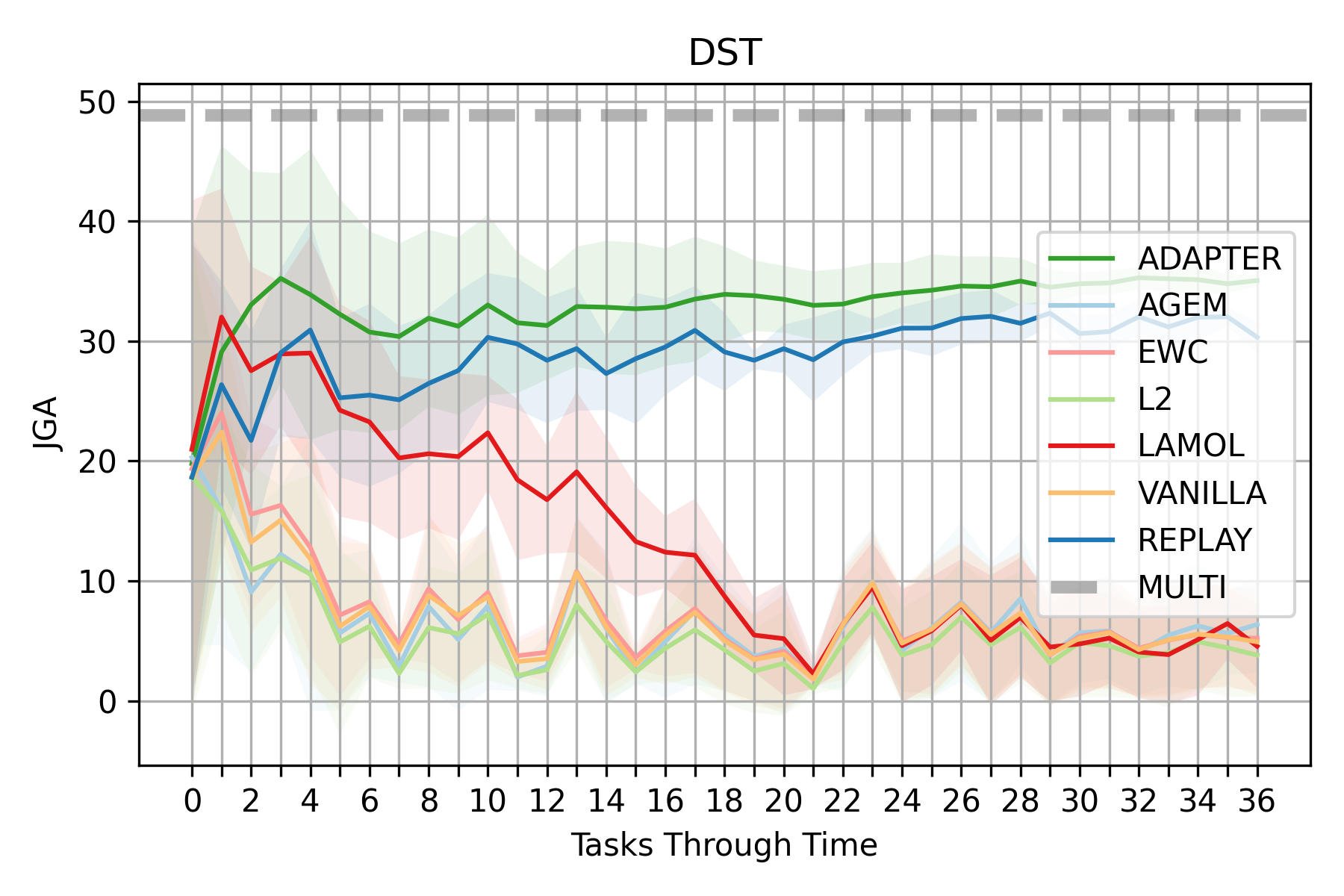}
    \includegraphics[width=\linewidth]{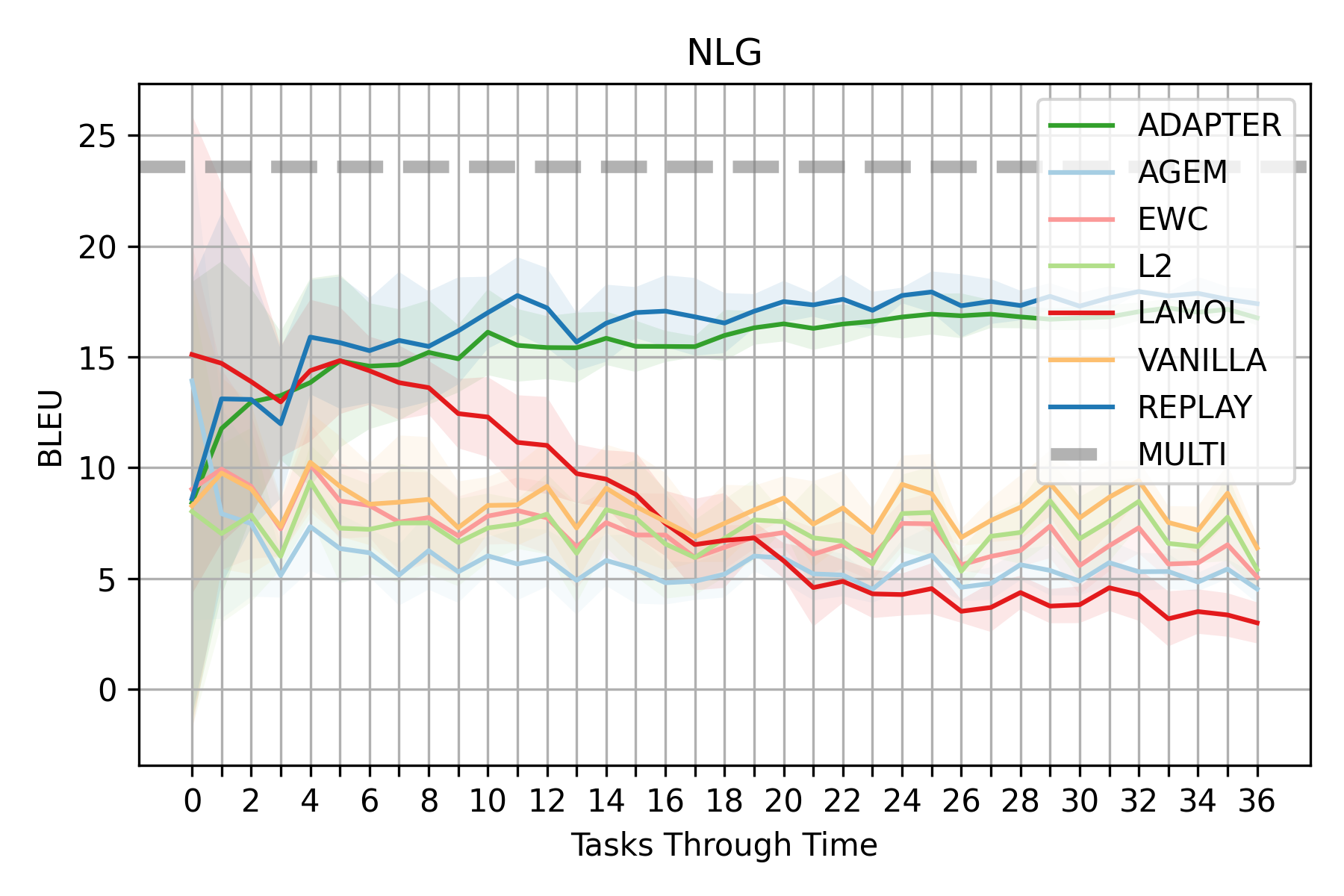}
    \includegraphics[width=\linewidth]{E2E_EER.png}
    \caption{Avg. Metric for the response generation, i.e., the BLUE and EER. }
    \label{fig:E2E2}
\end{figure}

\begin{figure}[t]
    \centering
    \includegraphics[width=\linewidth]{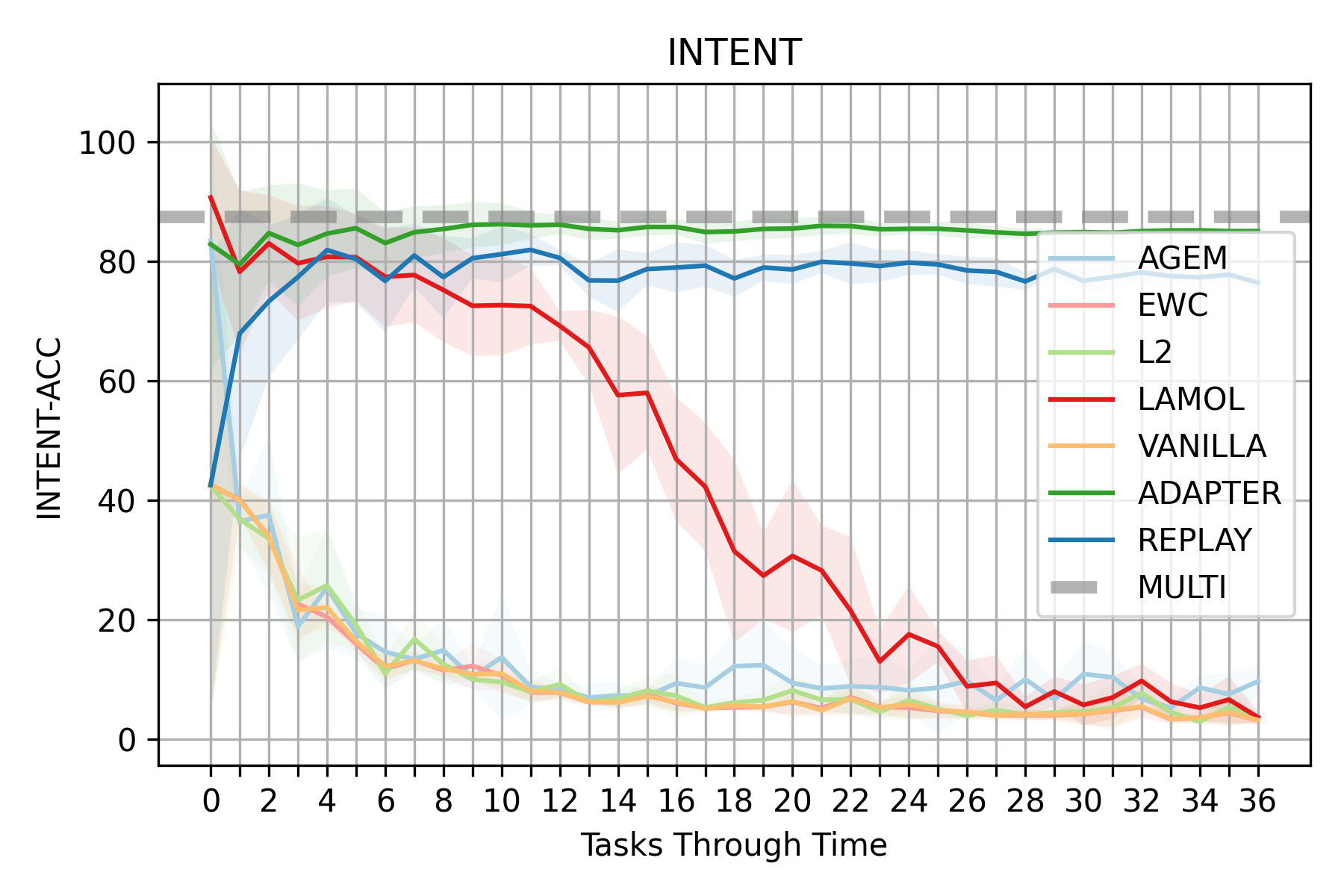}
    \includegraphics[width=\linewidth]{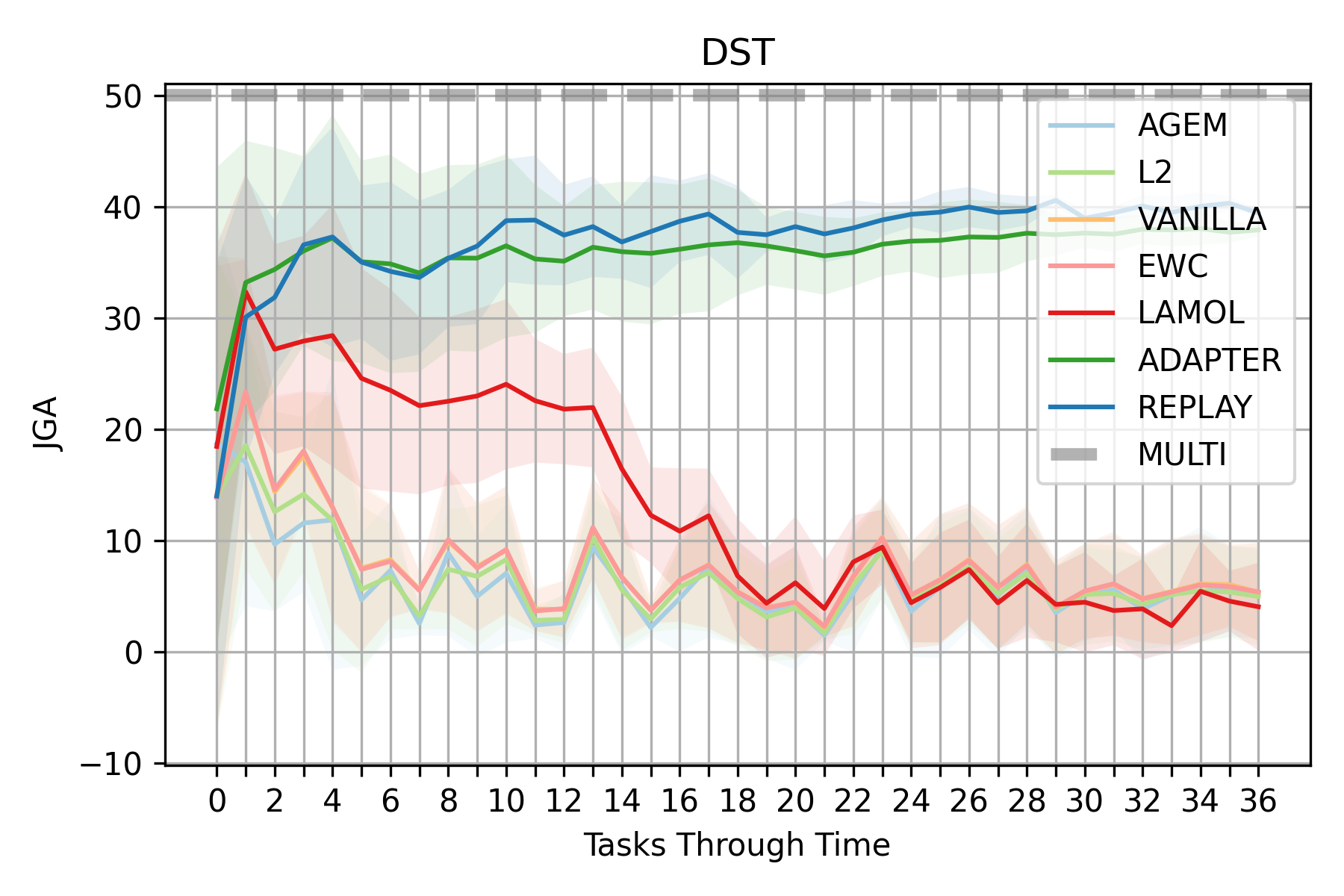}
    \includegraphics[width=\linewidth]{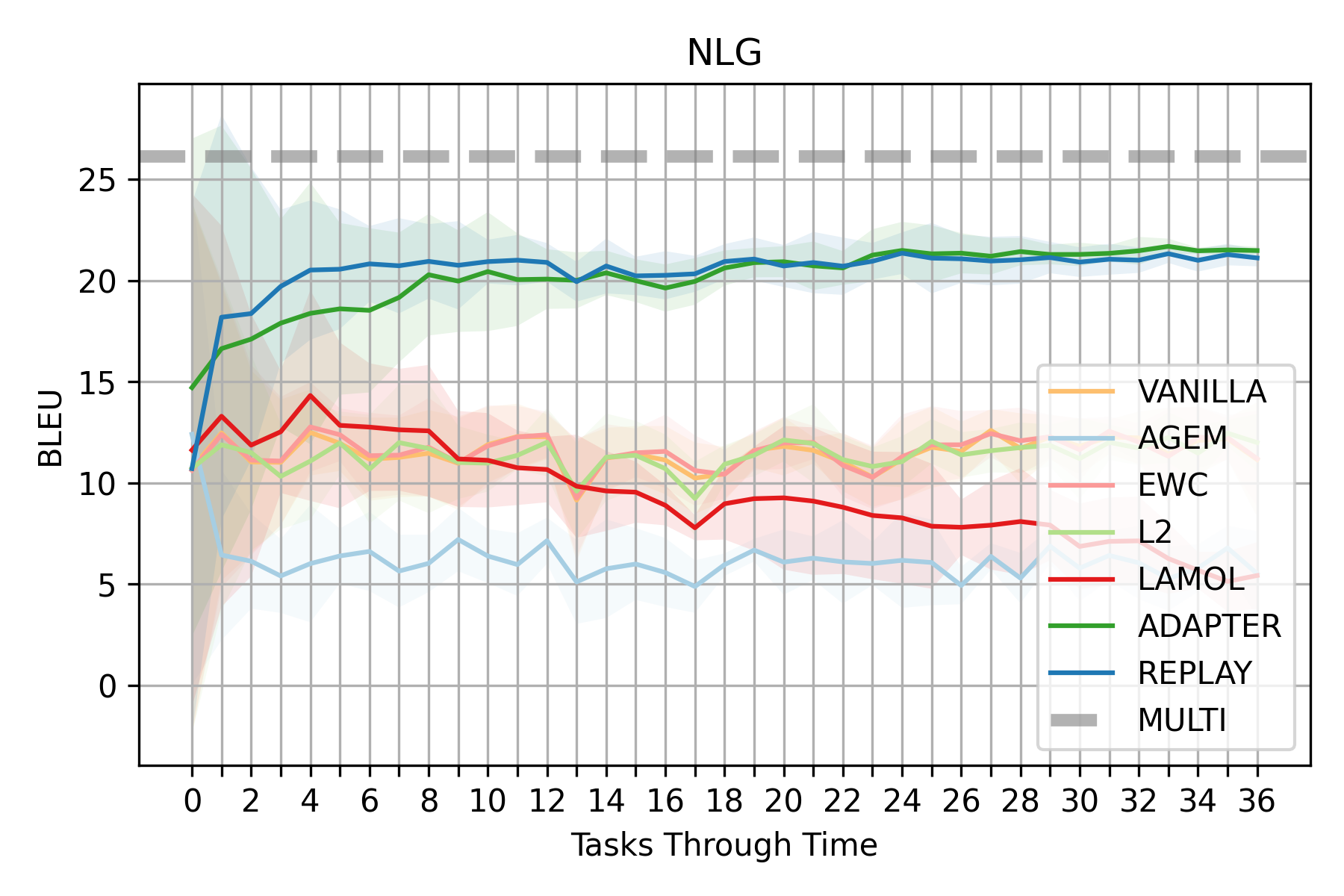}
     \includegraphics[width=\linewidth]{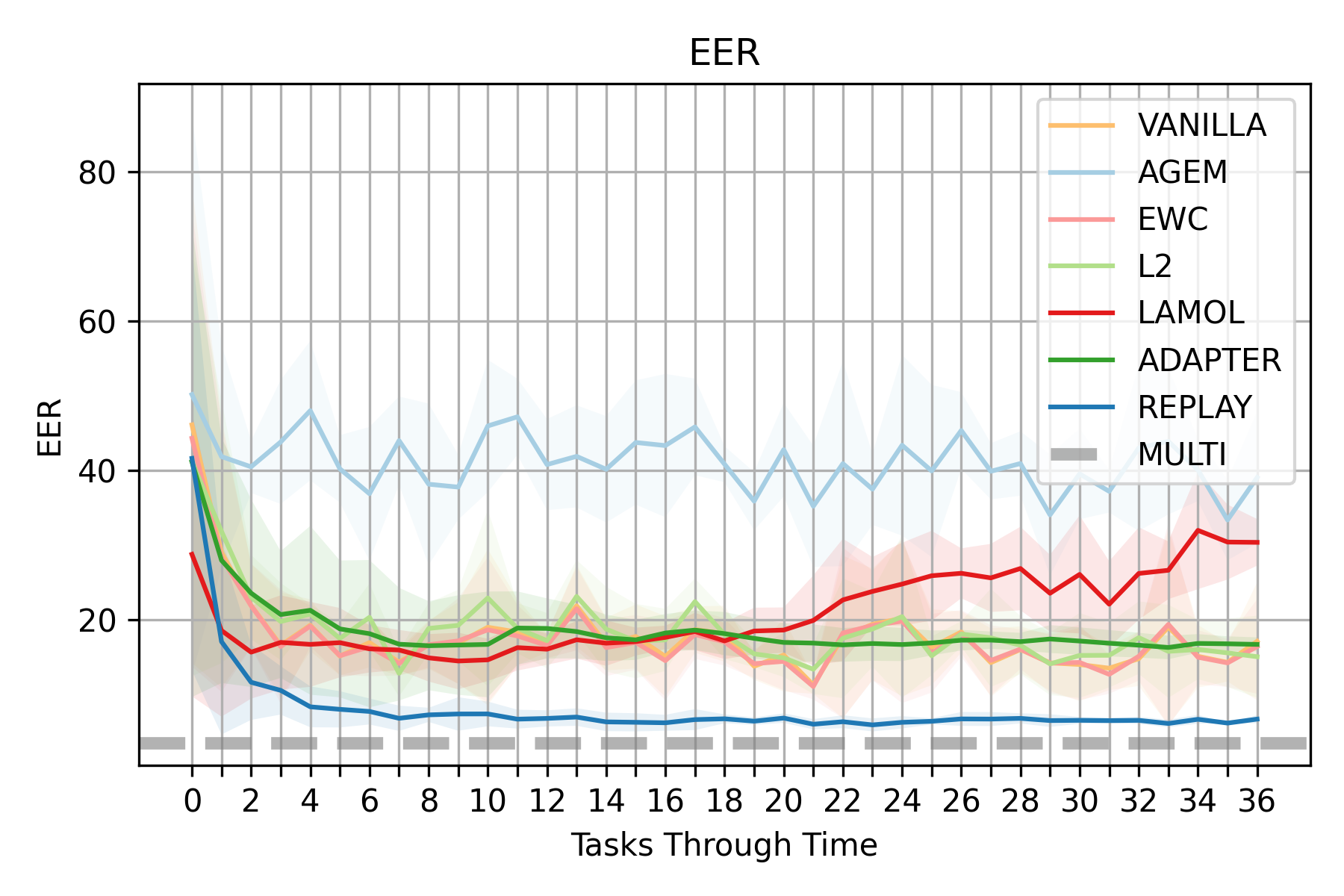}
    \caption{Modularized results}
    \label{fig:Module}
\end{figure}

\begin{figure}[t]
    \centering
    \includegraphics[width=\linewidth]{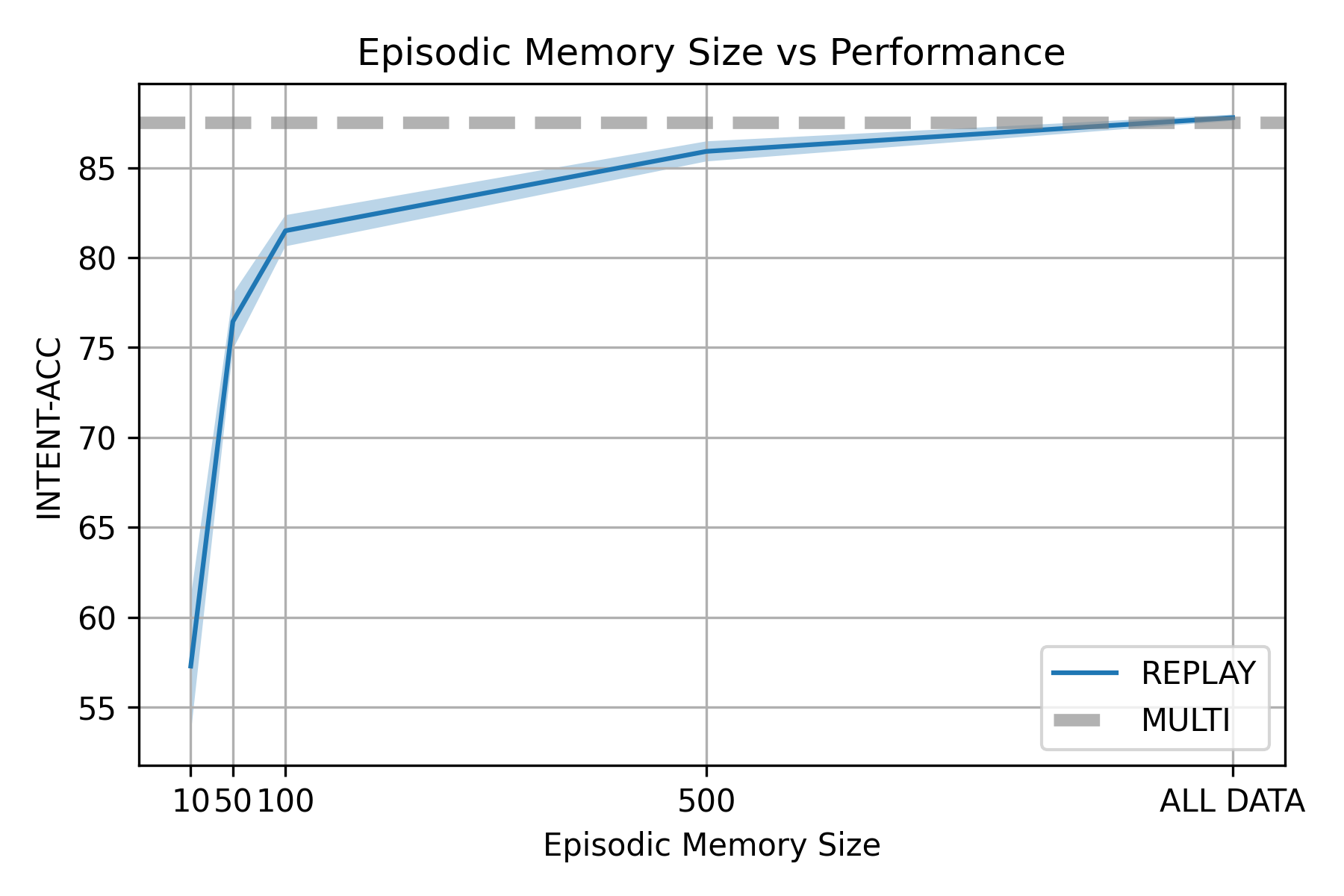}
    \caption{Ablation study on the size of the episodic-memory vs the intent accuracy. }
    \label{fig:intentmem}
\end{figure}

\begin{figure}[t]
    \centering
    \includegraphics[width=\linewidth]{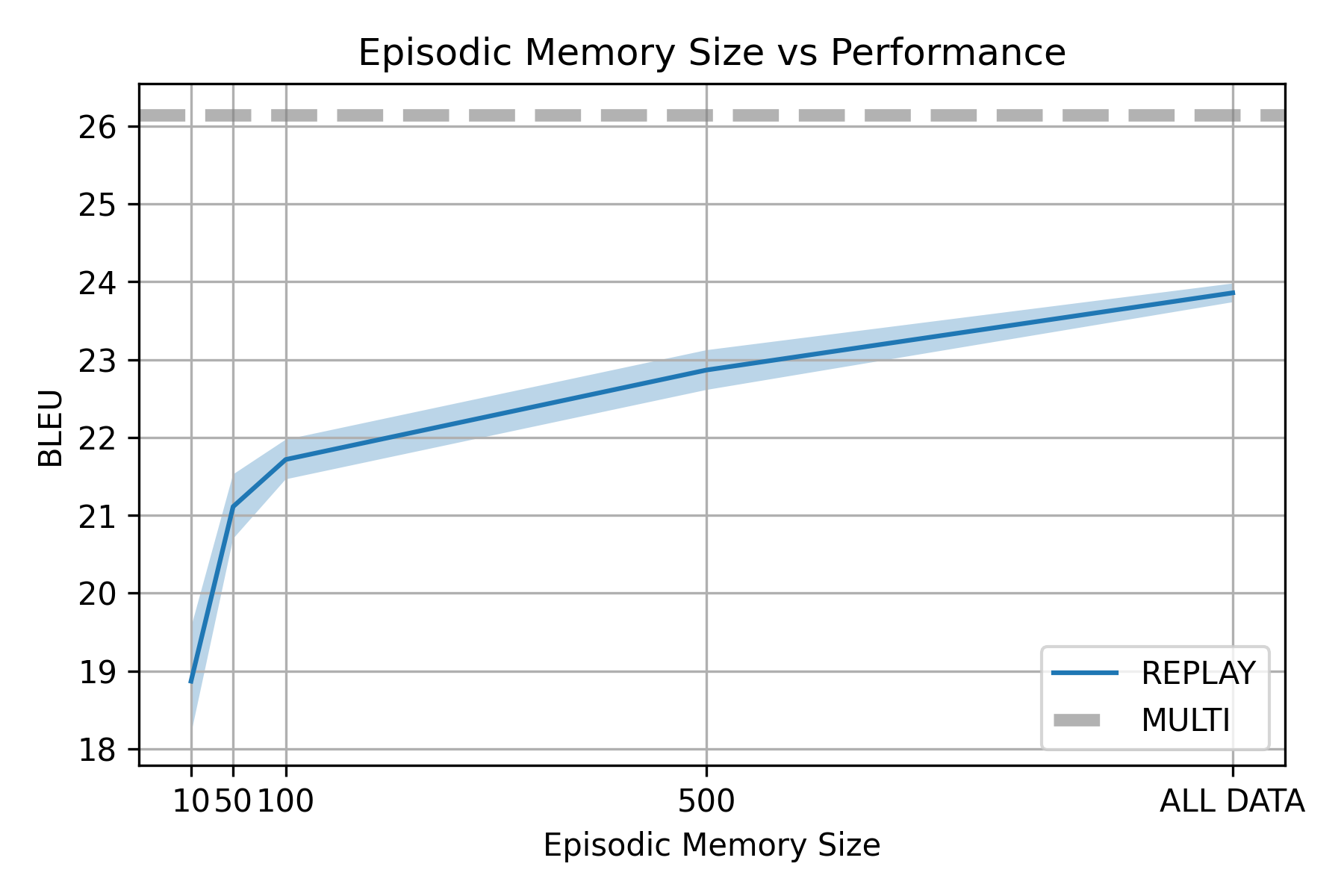}
    \includegraphics[width=\linewidth]{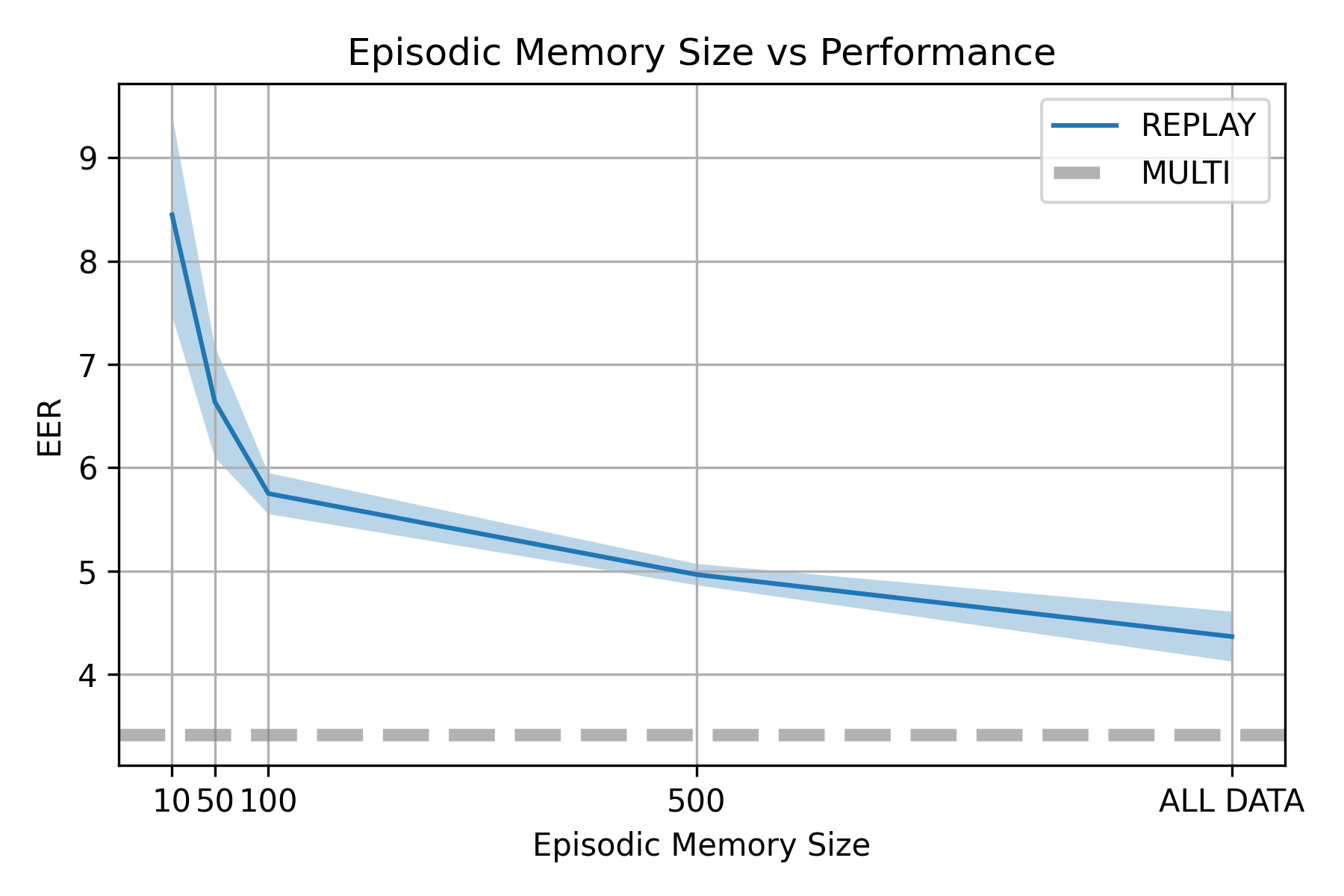}
    \caption{Ablation study on the size of the episodic-memory vs BLEU/EER. }
    \label{fig:BLUEEER}
\end{figure}


\begin{table*}[t]
\centering
\begin{tabular}{r|cc|c|c|cc}
                                     & \multicolumn{1}{l}{} & \multicolumn{1}{l|}{} & \textbf{INTENT}              & \textbf{DST}            & \multicolumn{2}{c}{\textbf{NLG}}                     \\ \hline
\multicolumn{1}{c|}{\textbf{Method}} & \textbf{+Parm.}      & \textbf{Mem.}         & \textit{Accuracy $\uparrow$} & \textit{JGA $\uparrow$} & \textit{EER $\downarrow$} & \textit{BLEU $\uparrow$} \\ \hline
{\ul \textit{VANILLA}}               & -           & $\emptyset$           & 3.27  $\pm$ 0.3              & 5.34  $\pm$ 4.4         & 14.81 $\pm$ 7.7           & 11.06 $\pm$ 2.9          \\
{\ul \textit{L2}}                    & $|\theta|$  & $\emptyset$           & 3.52  $\pm$ 0.7              & 4.95  $\pm$ 4.4         & 12.93 $\pm$ 5.5           & 11.99 $\pm$ 1.4          \\
{\ul \textit{EWC}}                   & $2|\theta|$ & $\emptyset$           & 3.21  $\pm$ 0.3              & 5.36  $\pm$ 4.3         & 14.2 $\pm$ 6.2            & 11.19 $\pm$ 2.4          \\
{\ul \textit{AGEM}}                  & -           & $t|M|$                & 9.74  $\pm$ 2.6              & 5.17  $\pm$ 4.0         & 34.2 $\pm$ 8.6            & 5.51 $\pm$ 2.1           \\
{\ul \textit{LAMOL}}                 & -           & $\emptyset$           & 3.73  $\pm$ 1.0              & 4.03  $\pm$ 3.9         & 29.61 $\pm$ 3.1           & 5.42 $\pm$ 1.7           \\
{\ul \textit{REPLAY}}                & -           & $t|M|$                & 76.45 $\pm$ 1.5              & \textbf{39.42} $\pm$ 0.2         & \textbf{4.95} $\pm$ 1.5            & \textbf{21.72} $\pm$ 0.3          \\
{\ul \textit{ADAPT}}                 & $t|\mu|$    & $\emptyset$           & \textbf{85.05} $\pm$ 0.6              & 37.9  $\pm$ 0.6         & 14.36 $\pm$ 0.7           & 21.48 $\pm$ 0.2          \\ \hline
{\ul \textit{MULTI}}                 & -           & -                     & 87.50 $\pm$ 0.2              & 50.04 $\pm$ 0.1         & 2.84 $\pm$ 0.2            & 26.15 $\pm$ 0.2          \\ \hline
\end{tabular}
    \caption{Results Modularized}
    \label{tab:modularized_results}
\end{table*}

\begin{table*}[t]
\centering
\begin{tabular}{c|c|c|cc}
\hline
                & \textbf{INTENT}              & \textbf{DST}           & \multicolumn{2}{c}{\textbf{NLG}}                    \\ \hline
$|\mathcal{M}|$ & \textit{Accuracy $\uparrow$} & \textit{JGA$\uparrow$} & \textit{EER$\downarrow$} & \textit{BLEU $\uparrow$} \\ \hline
10              & 57.286$\pm$3.80              & 26.63$\pm$1.26         & 8.44$\pm$0.97            & 18.86$\pm$0.68           \\
50              & 76.446$\pm$1.55              & 39.41$\pm$0.28         & 6.63$\pm$0.53            & 21.11$\pm$0.41           \\
100             & 81.496$\pm$0.86              & 43.13$\pm$0.31         & 5.75$\pm$0.19            & 21.71$\pm$0.25           \\
500             & 85.91$\pm$0.55               & 48.22$\pm$0.53         & 4.96$\pm$0.10            & 22.86$\pm$0.25           \\
ALL             & 87.784$\pm$0.16              & 49.97$\pm$0.46         & 4.36$\pm$0.24            & 23.85$\pm$0.12           \\ \hline
MULTI           & 87.5$\pm$0.1                 & 50.04 $\pm$ 0.6        & 3.42$\pm$0.1             & 26.15$\pm$0.1            \\ \hline
\end{tabular}
    \caption{Ablation study over episodic memory size $|\mathcal{M}|$. In the table $|\mathcal{M}|$ represent the number of samples per task kept in memory. }
    \label{tab:ablationmem}
\end{table*}

\begin{table*}[t]
\begin{tabular}{r|ccc|ccc|ccc}
\hline
\multicolumn{1}{r}{\textbf{Domains}} & \multicolumn{3}{c}{\textbf{DST-INTENT}}       & \multicolumn{3}{c}{\textbf{NLG}}              & \multicolumn{3}{c}{\textbf{End-to-End}}       \\ \hline
\multicolumn{1}{l}{}                 & \textit{Train} & \textit{Dev} & \textit{Test} & \textit{Train} & \textit{Dev} & \textit{Test} & \textit{Train} & \textit{Dev} & \textit{Test} \\ \hline
{\ul \textit{TM19 movie}}            & 4733           & 584          & 500           & 3010           & 366          & 341           & 12766          & 1632         & 1481          \\
{\ul \textit{TM19 auto}}             & 3897           & 448          & 522           & 2128           & 223          & 283           & 10918          & 1248         & 1443          \\
{\ul \textit{TM19 restaurant}}       & 4434           & 568          & 561           & 2582           & 330          & 333           & 12862          & 1669         & 1630          \\
{\ul \textit{TM19 pizza}}            & 2883           & 381          & 359           & 1326           & 171          & 171           & 8720           & 1145         & 1083          \\
{\ul \textit{TM19 uber}}             & 4378           & 535          & 525           & 2418           & 290          & 278           & 11331          & 1362         & 1361          \\
{\ul \textit{TM19 coffee}}           & 2591           & 302          & 335           & 1381           & 151          & 184           & 7429           & 894          & 936           \\
{\ul \textit{TM20 flight}}           & 15868          & 1974         & 1940          & 10148          & 1272         & 1245          & 36778          & 4579         & 4569          \\
{\ul \textit{TM20 food-ordering}}    & 3404           & 411          & 431           & 2394           & 277          & 287           & 7838           & 941          & 986           \\
{\ul \textit{TM20 hotel}}            & 15029          & 1908         & 1960          & 6590           & 842          & 869           & 35022          & 4400         & 4532          \\
{\ul \textit{TM20 music}}            & 5917           & 764          & 769           & 4196           & 537          & 523           & 13723          & 1773         & 1787          \\
{\ul \textit{TM20 restaurant}}       & 13738          & 1761         & 1691          & 8356           & 1063         & 994           & 34560          & 4398         & 4297          \\
{\ul \textit{TM20 sport}}            & 13072          & 1668         & 1654          & 12044          & 1553         & 1542          & 29391          & 3765         & 3723          \\
{\ul \textit{TM20 movie}}            & 13221          & 1703         & 1567          & 9406           & 1203         & 1093          & 32423          & 4158         & 3881          \\
{\ul \textit{MWOZ taxi}}             & 1239           & 234          & 194           & 402            & 71           & 56            & 2478           & 468          & 388           \\
{\ul \textit{MWOZ train}}            & 1452           & 158          & 160           & 563            & 63           & 59            & 2905           & 316          & 320           \\
{\ul \textit{MWOZ restaurant}}       & 5227           & 243          & 281           & 3333           & 141          & 177           & 10461          & 486          & 563           \\
{\ul \textit{MWOZ hotel}}            & 2798           & 289          & 385           & 1924           & 194          & 258           & 5602           & 579          & 771           \\
{\ul \textit{MWOZ attraction}}       & 484            & 43           & 42            & 295            & 27           & 26            & 975            & 86           & 85            \\
{\ul \textit{sgd restaurants}}       & 2686           & 278          & 616           & 1720           & 166          & 386           & 5756           & 606          & 1354          \\
{\ul \textit{sgd media}}             & 1411           & 230          & 458           & 988            & 167          & 324           & 3114           & 502          & 1005          \\
{\ul \textit{sgd events}}            & 4881           & 598          & 989           & 3241           & 389          & 590           & 10555          & 1317         & 2197          \\
{\ul \textit{sgd music}}             & 1892           & 275          & 556           & 1506           & 224          & 464           & 4040           & 597          & 1215          \\
{\ul \textit{sgd movies}}            & 1665           & 181          & 52            & 996            & 114          & 44            & 3760           & 420          & 126           \\
{\ul \textit{sgd flights}}           & 4766           & 1041         & 1756          & 2571           & 627          & 982           & 10429          & 2244         & 3833          \\
{\ul \textit{sgd ridesharing}}       & 652            & 85           & 187           & 377            & 48           & 107           & 1448           & 188          & 418           \\
{\ul \textit{sgd rentalcars}}        & 1510           & 250          & 469           & 865            & 153          & 280           & 3277           & 538          & 1009          \\
{\ul \textit{sgd buses}}             & 1862           & 331          & 653           & 1102           & 218          & 412           & 4050           & 709          & 1393          \\
{\ul \textit{sgd hotels}}            & 3237           & 394          & 948           & 1997           & 243          & 597           & 6983           & 858          & 2053          \\
{\ul \textit{sgd services}}          & 3328           & 360          & 926           & 2225           & 230          & 611           & 7262           & 803          & 2016          \\
{\ul \textit{sgd homes}}             & 2098           & 170          & 533           & 1312           & 96           & 338           & 4519           & 394          & 1158          \\
{\ul \textit{sgd banks}}             & 1188           & 139          & 293           & 723            & 84           & 181           & 2599           & 319          & 667           \\
{\ul \textit{sgd calendar}}          & 592            & 115          & 236           & 397            & 65           & 133           & 1313           & 246          & 501           \\
{\ul \textit{sgd alarm}}             & 212            & 34           & 91            & 221            & 30           & 74            & 580            & 82           & 198           \\
{\ul \textit{sgd weather}}           & 196            & 32           & 80            & 123            & 23           & 59            & 433            & 70           & 169           \\
{\ul \textit{sgd travel}}            & 186            & 23           & 48            & 121            & 14           & 30            & 420            & 53           & 106           \\
{\ul \textit{sgd payment}}           & 227            & 21           & 51            & 143            & 14           & 32            & 497            & 44           & 113           \\
{\ul \textit{sgd trains}}            & 300            & 73           & 128           & 149            & 43           & 66            & 668            & 158          & 274           \\ \hline
\multicolumn{1}{r}{\textbf{Total}}                 & 147254         & 18604        & 22946         & 93273          & 11722        & 14429         & 347885         & 44047        & 53641         \\ \hline
\end{tabular}
\caption{All data-samples used in the experiments}
\label{tab:all_data}
\end{table*}

\begin{figure*}
    \centering
    \includegraphics[width=\linewidth]{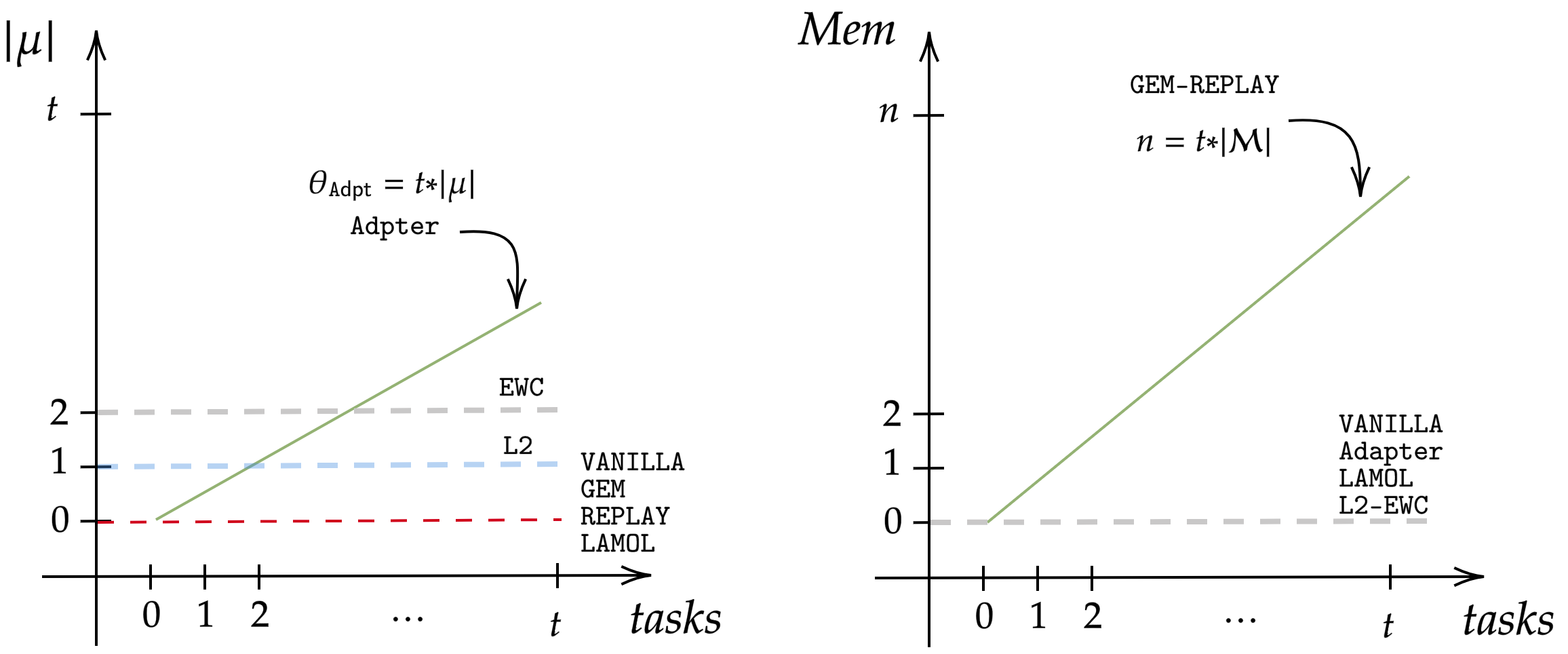}
    \caption{No Free Lunch in CL. This plot the trade-off between number of parameters added per task and the size of the episodic memory $\mathcal{M}$.}
    \label{fig:freelunch}
\end{figure*}

\end{document}